\documentclass[conference]{IEEEtran}
\usepackage{cite}
\usepackage{amsmath,amssymb,amsfonts}
\usepackage{algorithmic}
\usepackage{graphicx}
\usepackage{textcomp}
\usepackage{xcolor}
\usepackage{fancyhdr}
\usepackage[hyphens]{url}

\def\BibTeX{{\rm B\kern-.05em{\sc i\kern-.025em b}\kern-.08em
    T\kern-.1667em\lower.7ex\hbox{E}\kern-.125emX}}

\fancypagestyle{firstpage}{
  \fancyhf{}

  \fancyfoot[C]{\thepage}
}

\usepackage{tikz}
\usepackage{xcolor}

\usepackage{amsmath}
\usepackage{amssymb}

\usepackage{pgfplots}
\pgfplotsset{compat=1.16}

\usepackage{subcaption}
\usepackage{wrapfig}
\usepackage{threeparttable}

\renewcommand{\baselinestretch}{0.975}

\title{MILR: Mathematically Induced Layer Recovery for Plaintext Space Error Correction of CNNs} 
 \author{
Jonathan S. Ponader \\
  University of Central Florida\\
  Orlando, FL 32816 \\
  \texttt{JPonader@knights.ucf.edu} \\
\and
  Sandip Kundu \\
  University of Massachusetts \\
  Amherst, MA 01003\\
  \texttt{Kundu@ecs.umass.edu}\\
\and
  Yan Solihin \\
  University of Central Florida\\
  Orlando, FL 32816 \\
  \texttt{Yan.Solihin@ucf.edu} \\
 }

\begin{document}
\maketitle
\thispagestyle{firstpage}
\pagestyle{plain}


\begin{abstract}
  The increased use of Convolutional Neural Networks (CNN) in mission critical systems has increased the need for robust and resilient networks in the face of both naturally occurring faults as well as security attacks. The lack of robustness and resiliency can lead to unreliable inference results. Current methods that address CNN robustness require hardware modification, network modification, or network duplication. This paper proposes \emph{ MILR } a software based CNN error detection and error correction system that enables self-healing of the network from single and multi bit errors. The self-healing capabilities are based on mathematical relationships between the inputs,outputs, and parameters(weights) of a layers, exploiting these relationships allow the recovery of erroneous parameters (weights) throughout a layer and the network. MILR is suitable for plaintext-space error correction (PSEC) given its ability to correct whole-weight and even whole-layer errors in CNNs. 
\end{abstract}
\section{Introduction}

Artificial Intelligence (AI) is increasingly used or considered in mission critical systems, whether used in the cloud or in the field (e.g. autonomous cars, industrial control systems)~\cite{businesswire_2019}. Such systems must remain robust and resilient in the face of naturally occurring faults. In particular, for AI system relying on neural networks (NN), soft memory errors may result in corruption of memory values representing weights in the NN. Memory failure rate remain a big concern in the future as DRAM scaling is pushing the limit, while new memory technologies suffer from higher expected error rates. These memory faults can be caused by many environmental factors such as temperature, radiation, wear-out of hardware, and device specific issue such as resistance drift~\cite{ResistanceDrift}. While some memory faults may be masked out, others may cause {\em silent data corruption} (SDC) that lead to unintended consequences such as misclassification~\cite{Qin2017RobustnessON,Li2017UnderstandingEP,Stevenson1990SensitivityOF}.

\begin{figure}[htbp]
    \centering
    \includegraphics[scale=0.4]{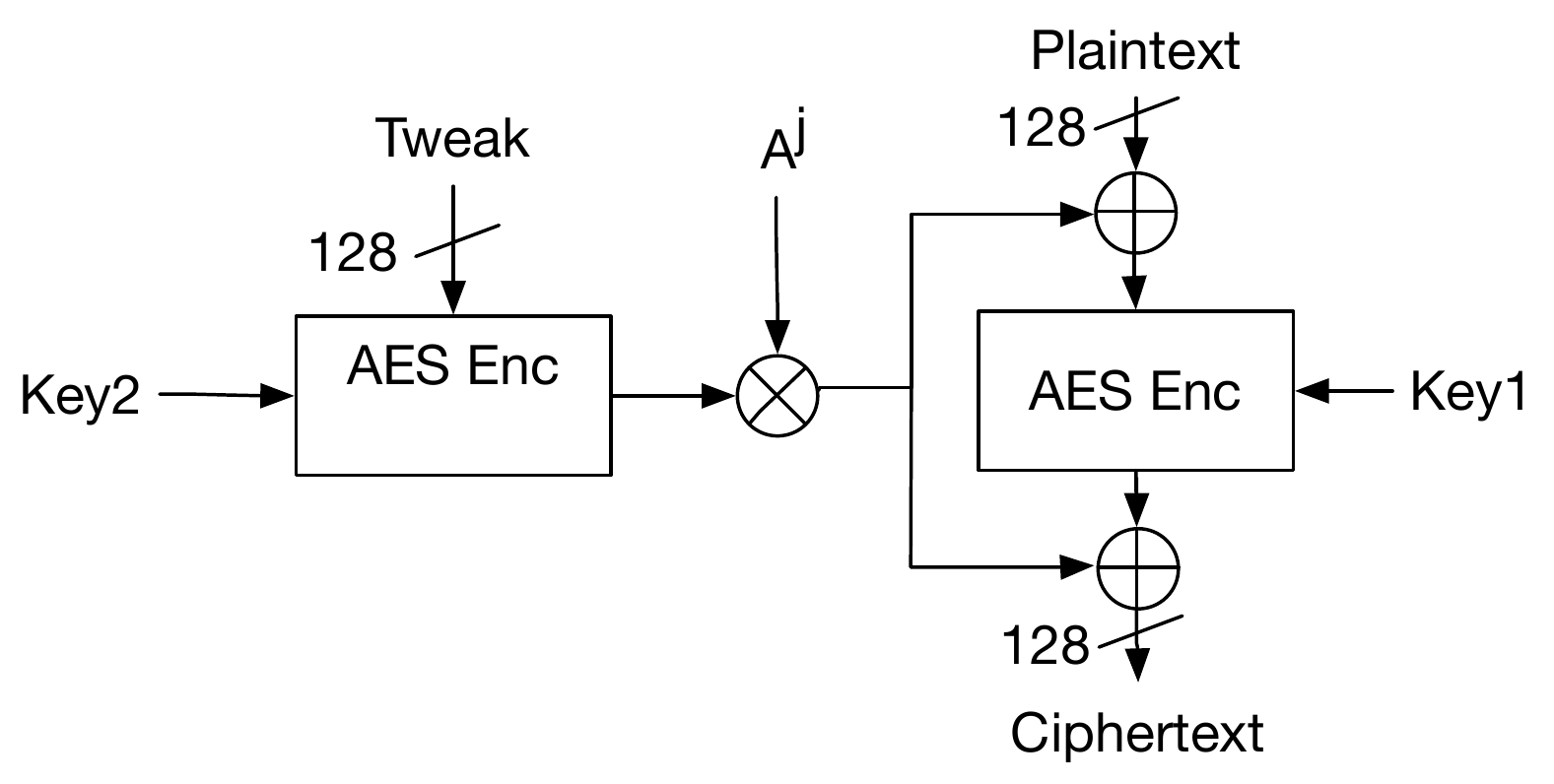}
    \caption{AES-XTS encryption mode used for memory encryption in MKTME.}
    \label{fig:aes-xts}
\end{figure}

AI software also represent a highly valuable intellectual property that is high valued target for thefts~\cite{web:googletheft}. When deployed in the cloud, its security can be improved by utilizing encrypted isolated environment, such as running it in an encrypted virtual machine (VM). Both AMD secure encrypted virtualization (SEV)~\cite{AMDSEV} and Intel multi-key total memory encryption (MKTME)~\cite{Intel-MKTME}\footnote{Intel also provides secure enclave environment through SGX. However, full AI software is unlikely to run in an SGX enclave due its memory limited to 128MB.}  keep each VM encrypted using a unique key, preventing other VMs or the hypervisor from learning about plaintext of data processed by the AI software. MKTME relies on AES-XTS mode~\cite{Intel-MKTME} for encrypting memory (Figure~\ref{fig:aes-xts}).

NN resiliency depends on (1) the ability to detect errors and (2) the ability to recover from errors without the involvement of external intervention ({\em self-healing}). In this study, we limit our concerns to self healing from soft memory errors. Current solutions rely on error correcting code (ECC) for detecting and correcting errors, for example a popular Hamming code SECDED detects two bit errors and corrects one bit error in a word. 

For NN software running on encrypted VM, we distinguish between the memory that contains  encrypted data ({\em ciphertext space}) and plaintext data ({\em plaintext space}). We point out that ECC schemes are appropriate for the ciphertext space, but not so for the  plaintext space. The primary assumptions ECC relies on are that (1) errors are distributed randomly on a bit level, and (2) the probability of multi-bit error in a word is low. With these assumptions, we only need to correct a small number of bit errors per word, and protect each word with a separate ECC. Unfortunately, these assumptions do not apply in the plaintext space. An uncorrected bit error in the ciphertext of a word translates to many-bit error in the plaintext after decryption in AES-XTS mode. Furthermore, the error is no longer randomly distributed in the plaintext space; they are concentrated in  bits that belong to an encryption word. As a result, {\em a different and more powerful error correction is needed in the plaintext space: one that can deal with many bit errors in entire encryption words}.

In contrast to prior ECC schemes that focus on ciphertext space, we focus on a new problem of how to provide {\em plaintext space error correction} (PSEC), with a goal of providing self-healing of NN systems. We limit our scope to convolutional neural network (CNN) systems, but the problem and approach likely extend to other NNs. Fundamentally, it is difficult to correct many-bit errors in the plaintext space because error correction capability depends on code that capture information redundantly, hence the amount of redundancy required increases with the amount of desired correctable errors. However, we make a novel observation that redundancy of information that naturally occurs in CNN systems can be leveraged for PSEC. We propose MILR (which stands for Mathematically-Induced Layer Recovery), a PSEC software solution for CNNs. MILR relies on a key observation that the input, output and parameters (weights) of a CNN layer are algebraically related, and that in many cases, knowing two of them allow the recovery of the third. We show that MILR can correct not just multi-bit errors in a word, it can detect and correct errors affecting entire weights, and in many cases, a entire NN layer, in addition to the regular random bit error. 

We envision MILR can have several novel uses. First, for CNN software running on encrypted VM, MILR could enhance ECC-protected memory by detecting and correcting errors in the plaintext space that escape ECC. With this use, the resiliency of CNN to errors increases substantially, making it suitable for self healing of systems that need high classification accuracy despite high memory error rates. Second, MILR can be used in lieu of ECC, in some embedded systems where ECC use is prohibitive. MILR is a software solution, so it can be applied selectively and only as needed. In general, MILR achieves higher effectiveness in correcting errors than SECDED ECC. Third, MILR can also be used to self heal from security attacks that involve memory corruption in the plaintext space. For example, an attacker, exploiting software vulnerability, may cause an overwrite to a targeted weight in a NN to force classification error. MILR can detect weights that have been modified and restored them.  


MILR was evaluated with three CNN networks in an error simulator, injecting the network with random bit flips with varying Raw Bit Error Rate (RBER)\cite{ElementsIT}, randomly flipping all bits in a weight at varying error rates and overwriting all weights in a layer with random values. These test simulate both soft memory errors and security attacks. MILR was able to increase the robustness of these networks, enabling them to operate normally even after being subjected to these modifications. We found that MILR corrects whole weight or even whole layer errors in plaintext space where ECC cannot. Even in the ciphertext space, MILR can tolerate higher bit error rates than SECDED ECC. Finally, we showed how   availability-accuracy trade off curves can be derived for MILR that help users select the most mission-appropriate design. 
\section{Related Works}
The fault tolerance of neural networks has been studied at depth, both from a robustness standpoint \cite{Qin2017RobustnessON,Li2017UnderstandingEP,Stevenson1990SensitivityOF} as well as a security standpoint \cite{Rakin2019BitFlipAC, Liu2017FaultIA} attempting to exploit the lack of robustness in a network. All of these works have shown that bit-flips can have a major impact on the networks performance. Li et al.~\cite{Li2017UnderstandingEP} shows that the impact of bit flip errors on NN accuracy depends on the type of network, data types (e.g. float or float16), and bit position. While soft errors may be distributed randomly, security attacks may rely on targeting certain bits that are extremely impactful on NN performance. For example, Rakin et al.\cite{Rakin2019BitFlipAC} showed that in the  ResNet-18 network with 93 million bits of parameters it only took 13 bit flips to degrade the accuracy from \( 69.8\% \)accuracy to \( 0.01\%\).

Many solutions have been proposed to address bit flips in neural networks, both for detecting and for recovering from them. All such works assume non-encrypted VM. ECC has long been used for detecting and correcting memory errors in an application agnostic manner \cite{SridharanVilas2015MemoryEI}. Guan et al. \cite{Guan2019InPlaceZM} proposed a application specific ECC approach for convolutional neural networks, storing 1 parity bit within a quantized 8-bit weight, combining 8 together to make a 64 bit word with an 8 bit ECC parity string. Both versions of ECC suffer from limited detection and recoverability leaving them susceptible to multi-bit errors. Li et al.\cite{Li2017UnderstandingEP} proposed a symptom based error detector and selective latch hardening. The symptom based error detector works by detecting unusual values of variables, which are likely to cause SDC. The normal range of values are obtained from training, making it hard to add protection after deployment. The selective latch hardening reduces vulnerabilities of logic to soft errors. 

Triple Modular Redundancy (TMR) is another agnostic approach to recovery from both logic and memory errors. TMR works by running three copies of the same application and a majority vote for the results is used~\cite{Lyons1962TheUO}. Self-healing can be provided if the system allows the majority instances to update the NN of the minority instance. TMR is expensive as it requires tripling of computation and memory resources. Dual Modular Redundancy (DMR) is another variation of redundancy approach but is only able to detect errors but the lack of the third copy makes it difficult to figure out which instance is the erroneous one, preventing self healing. Phatak et al.~\cite{Phatak1995CompleteAP} proposed only duplicating parts of hidden layers in the network. Such an approach is less expensive than a standard TMR but does not provide complete fault tolerance. Qin et al. \cite{Qin2017RobustnessON} explored not recovery errors but setting erroneous weights to zero. This decreased the accuracy drop caused by bit-flips, but higher error rates still caused a significant loss in performance.
\section{MILR: Mathematically Induced Layer Recovery}
MILR works on the premise that there is an algebraic relationship between the input, output and parameters of each layer of a CNN. Suppose that an CNN layer \(F\) receives input \(X\) to produce output \(Y\) using parameters \(P\). Given the input and parameters, the output can be computed using forward pass (Equation \ref{eq:forward}). Given the output and parameters, the input can be computed using a backward pass (Equation \ref{eq:backward}), if the layer is invertible. Given the input and output, the parameters can be computed using parameter solving function \(R\) (Equation \ref{eq:parameter}). 
\begin{eqnarray}
 f(x,p) &=& y  \label{eq:forward} \\
 f^{-1}(y, p) &=& x  \label{eq:backward} \\
 R(x,y) &=& p \label{eq:parameter}
\end{eqnarray}

Exploiting the relationships in the above equations forms the foundation of MILR. Suppose that a pair of known-good ({\em golden}) input and output is stored in reliable and safe memory, while the parameters are placed in main memory that provides fast access but is prone to errors and attacks. Since outputs change if there are errors in parameters, to provide error detection, MILR utilizes a forward pass with the golden input, and compares its output with the golden output; a mismatch indicates parameter errors. To provide self-healing, the parameters are recomputed using the pair of golden input and output, and the recomputed parameters overwrite the erroneous parameters. Using such an approach, MILR can provide both error detection and self-healing. Note, however, while recovery phase is only needed when errors are detected, error detection phase must be scheduled before errors can be detected. If instant error detection is required, a different scheme should be used. To cater for the possibility of needing a different error detection scheme, we wish to keep the design of error detection and recovery separate, even though they still rely on the same principle of exploiting algebraic relationship between input and output. 

\begin{figure}[htbp]
    \centering
    \includegraphics[scale=0.32]{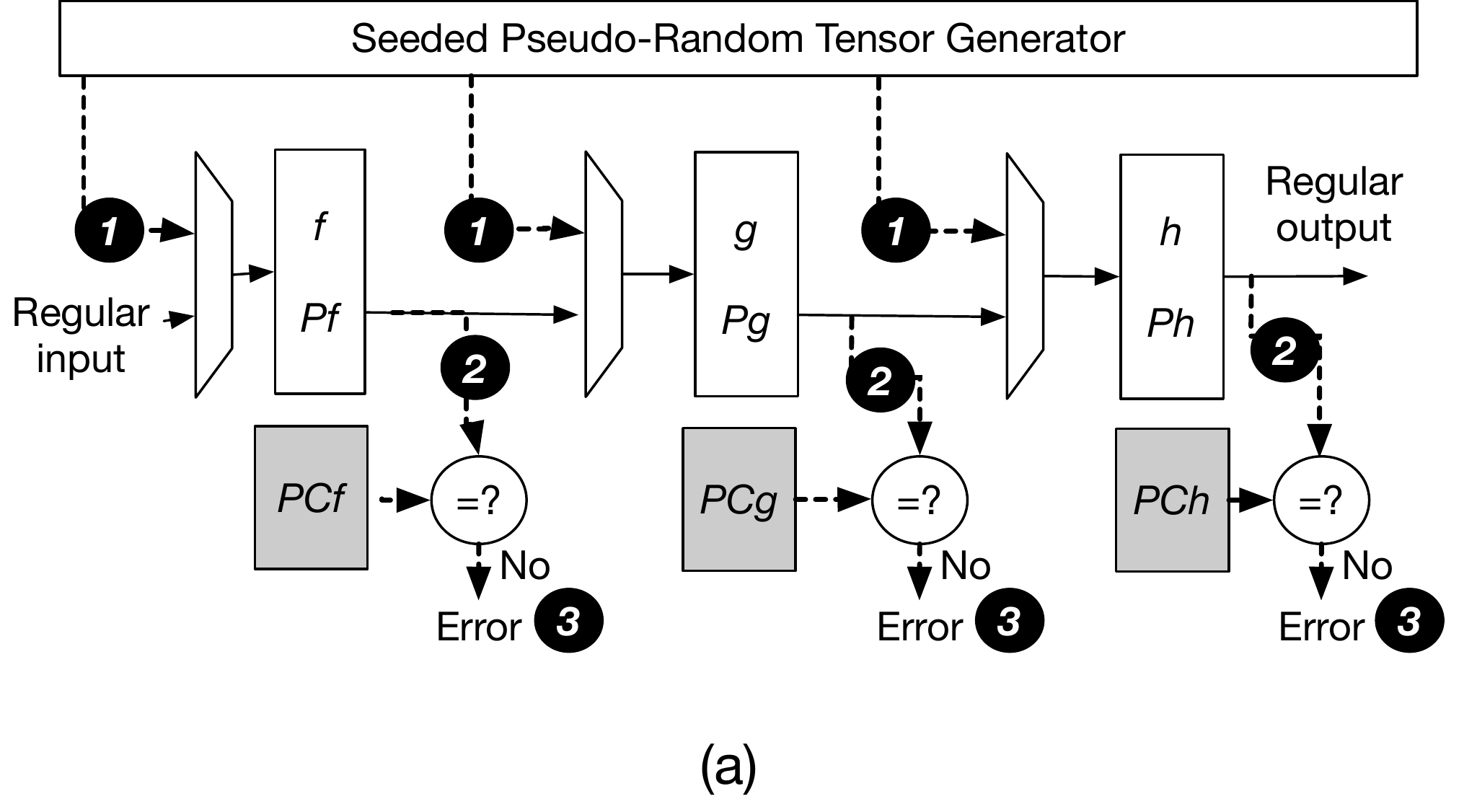}
    \caption{Illustration of MILR error detection.}
    \label{fig:milr-detection}
\end{figure}

Figure~\ref{fig:milr-detection} illustrates an example MILR error detection phase with three error-vulnerable layers \(f, g, h\), with \(P_f, P_g, P_h\) their respective parameters. Solid lines indicate normal execution flow, while dashed lines indicate error detection flow. Many layers in an NN exhibit a repeated behavior, using the same subset of parameters multiple times to produce multiple outputs. This behavior allows use to create something called a {\em partial checkpoint}. A partial checkpoint stores a single output value from each subset of the layers parameters, reducing the storage overhead while still allowing for erroneous layers to be identified. Tremendous savings occur using partial checkpoints: a partial checkpoint can be up to two orders of magnitude smaller than a full checkpoint for convolutional layers. Partial checkpoints,\(PC_f, PC_g, PC_h\) are stored  in error-resistant memory and correspond to the output of each layer given an input that is generated using pseudo-random number generator. By using pseudo-random number generator, we only need to memorize the initial seed, and the partial checkpoints to allow erroneous layers to be identified. To initiate error detection phase, input is constructed using pseudo random number generators \textcircled{1}, and each layer initiates a forward pass. The output of each layer is then compared against the partial checkpoint \textcircled{2}, and if they do not match, the layer is flagged as containing erroneous parameters \textcircled{3}.
\begin{figure}[htbp]
    \centering
    \includegraphics[scale=0.32]{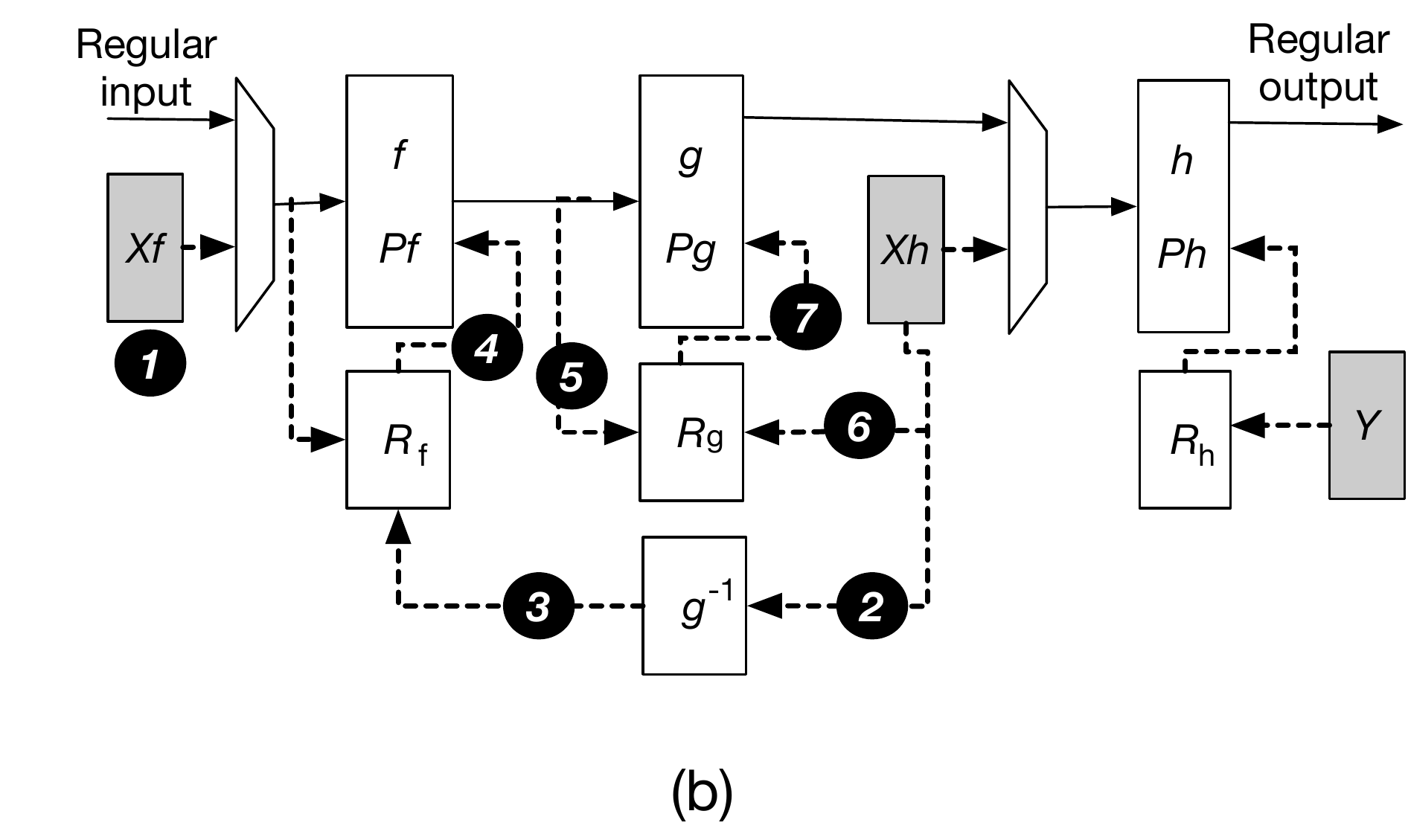}
    \caption{Illustration of MILR self-healing recovery}
    \label{fig:milr-recov}
\end{figure}

Figure~\ref{fig:milr-recov} illustrates an example MILR error recovery phase providing self-healing. Error recovery requires full checkpoints due to the need to recover the exact values of parameters, hence a challenge is the space overheads required to store checkpoints. To address the storage overhead, we observe three opportunities for removing checkpoints. First, we can skip keeping input checkpoint if a layer is invertible, since an output checkpoint can be used with a backward pass to calculate the input. In the figure, we store the golden checkpoints \(X_f, X_h\) as well as the golden output checkpoint \(Y\) in error-resistant memory, but skip \(X_g\) because \(g\) is invertible. Normal CNN inference flows along the solid lines, while error recovery flows along the dashed lines. When error recovery is needed, the identified erroneous layer needs a golden input/output pair as input to its parameter recovery function. Suppose that layer \(f\) is erroneous. Its parameter recovery function receives input \(X_f\) \textcircled{1}. \(X_h\) is input to the inverse of \(g\) \textcircled{2}, i.e. \(g^{-1}(X_h, P_g)\), generating an output needed for the parameter recovery function \(R_f\) \textcircled{3}. \(R_f\) then recovers the parameters, self-healing layer \(f\) \textcircled{3}. Hence, layer \(f\) is recovered even without checkpoint \(X_g\) because \(g\) is invertible. As another example, suppose that layer \(g\) is erroneous. In this case, its parameter recovery function receives the output of layer \(f\) \textcircled{5} and the checkpoint \(X_h\)  \textcircled{6} without a need for a backward pass. Then, the parameter recovery function \(R_g\) uses both as inputs to recover layer \(g\) parameters \textcircled{7}.

The second opportunity for removing an input checkpoint is when the preceding layers have no parameters. Since preceding layers have no parameters to recover, it does not need input/output pairs to recover. Hence, we can remove the input checkpoint. 

The third opportunity is when we can transform a non-invertible layer to an invertible layer. A layer may not be invertible when there is not enough information encoded in the layer, e.g. the number of equations in the system of equations is lower than the number of parameters to be solved. Such a layer can be made invertible by adding {\em dummy} inputs that are used only for error recovery. Dummy inputs do not need to be stored in the input checkpoint if we rely on pseudo-random number generators for such input, however the additional output values will need to be stored (dummy outputs).

Given that not every checkpoint is kept between two consecutive layers, the recoverability of a network is limited by the number of checkpoints.
To recover parameters of a layer, we need to do forward pass from the a preceding checkpoint and backward pass from a succeeding checkpoint, thus the system can only recover at most one layer in between two checkpoints, but any number of parameter errors in that layer can be recovered. This is substantially more powerful than traditional ECC which can only recover 1 bit error in one word, hence even a multi-bit error of a single parameter cannot be recovered, let alone all parameters in a single layer. Furthermore, if there are \(N\) checkpoints, we can recover up to \(N-1\) layers as long as there is at most one layer with errors in between each pair of checkpoints.  

MILR is divided into 3 distinct phases: the initialization, error detection, and the error recovery phases. In the initialization phase all the additional data needed for detection and recovery is calculated and stored, including seeds for pseudo-random number generators, partial and full checkpoints, and dummy outputs needed to make some non-invertible layers invertible. In choosing between using a checkpoint or dummy input/output, we chose the strategy that minimizes storage overheads. The initialization phase only runs once when neural network is started on a system. In the error detection phase the seeded pseudo-random number generator regenerates the known input for each layer and is used in a forward pass to generate an output. The output is compared against the partial checkpoint to test for a mismatch (indicating errors in the layer). A log of erroneous layers and the location of the nearest input and output checkpoints is created for use by the recovery phase. The error recovery phase uses this log to identify layers to be recovered. Using the current locations of the checkpoints, they are moved through the network with a forward/backward pass to the affected layer. Once the input and outputs are available to the affected layer, the parameter recovery function is called to recover and restore the erroneous parameters.

Partial and full checkpoints are not stored in the (DRAM) working memory which is subject to soft errors as well as memory-based security attacks. They can be stored in error-resistant mediums, such as the storage devices (SSD or HDD) or persistent memory (e.g. Intel DC persistent memory). This is because the storage subsystem or persistent memory are much denser and cheaper (by orders of magnitude) can be made very dependable through redundancy techniques such as RAID, and their slow access is tolerable because the checkpoints are only needed for the occasional error detection or recover.  
\section{Layer Specific Error Detection and Recovery}
While MILR may be general enough for other types of NNs, we apply MILR on CNNs that are composed of 4 major layer types: convolution, dense, pooling, and ReLu (Activation Layers)~\cite{Wu2017IntroductionTC}, including the bias and activation parts of both dense and convolution layers. However these parts will be handled as independent layers as each part has their own mathematical relationships between the input, output and parameters. 

\subsection{Dense Layers}
Dense layers use matrix multiplication to compute their output, with a 2D tensor of shape \((M, N)\) as the input. This tensor is multiplied with a 2D parameter tensor of shape \((N, P)\), producing a 2D tensor of shape \((M, P)\) as the output. In other words, \(A_{(M,N)} \times B_{(N,P)} = C_{(M,P)}\), where A represents input, B represents parameters, and C represents output. 

\paragraph{Backward Pass}
The backward pass of matrix multiplication can be yield input \(A\) by computing \(C \times B^{-1}\). For this to be done the input shape and the parameter shape must meet certain requirements. The parameter must be of size such that \(P \geq N\). If these requirements are not met, additional information will need to be stored to allow for the creation of enough equations for the system of equations. If \(P < N\), then the \(P\) dimension of parameters will be padded with \(\alpha\) {\em dummy parameters} such that \(P+\alpha \geq N\). In order to minimize the storage overhead due to the dummy parameters, the dummy parameters are a stream of pseudo-random numbers, hence only the seed needs to be stored, along with the output. Dummy parameters produce more equations that allow the a system of equations to be solvable. We also compare the cost of storing dummy parameter output vs. a checkpoint, as a checkpoint removes the need for a layer to be invertible, and choose the approach that incurs a lower cost. 

\paragraph{Parameter Solving}
Parameter solving works on the same principle as the backward pass. To solve for parameters, the input shape must satisfy \(M \geq N\) to create a solvable system of equations, allowing correct parameters to be recovered. If \(M < N\), then the input will need to be padded with dummy input along the \(M\) dimension such that \(M \geq N\). Just as in the backward pass, the padding relies on a seeded pseudo-random number generator, to avoid storing dummy input. 

\paragraph{Error Detection \;}
In the dense layer, each parameter column can be used multiple times on different input rows. This leads to the output of a single parameter column to appear multiple times in the output. Only storing one of these outputs per parameter column can therefore allow for error detection. As we have an output that is resultant of each parameter value (i.e. the partial checkpoint), it can identify if any parameter changes. 

\subsection{Convolution Layers}

Convolution layers are the foundation of CNNs. They work by shifting a filter (parameters) along the \(X\) and \(Y\) axis of the input, taking a weighted sum of the covered input  sub-region~\cite{Wu2017IntroductionTC}. The input into a convolution layer is a 3D tensor of shape \((M, M, Z)\), where \(Z\) is the number of channels. This input is then processed by \(Y\) filters, each of shape \((F,F,Z)\), represented by a 4D Tensor of shape \((F, F, Z, Y)\). Producing an output tensor of shape \((G, G, Y)\). The relationship between the input size and the output size can be expressed as \(G = ((M-F + 2P)/S)+1\), where \(S\) is the stride of the convolution and \(P\) is the padding added to the input. Assuming that the stride is 1, Equation \ref{eq:conv} is repeated for all filters \(0 < k < Y\), and all locations \(0 < i < G\)  and  \(0 < j < G\).
\begin{equation}
 Out_{i,j,k} = \sum_{f_1=0}^{F} \sum_{f_2=0}^{F} \sum_{z_1=0}^Z Filter_{f_1,f_2,z_1,k} \times In_{{f_1+i},{f_2+j},z_1}
 \label{eq:conv}
\end{equation}
\paragraph{Backward Pass \;}
The backward pass through the convolution layer is based on the observation that each filter looks at the same sub-region of the input, hence \(Y\) filters look at each sub-region of the input. This produces \(Y\) equations each representing the weighted sum of the sub-region, which can be used as a system of equations to recover the sub-region. In order to have enough equations to make the system solvable, \(Y \geq F^2Z\) is required.

If \( Y < F^2Z\), additional information is needed to be able to conduct the backward pass. For a convolution layer, we can generate more filters to create additional equations. As in dense layers, we rely on pseudo-random number generator to create additional filters so that only the seed and output need to be stored. Also, as in dense layers, we compare the storage overhead of additional filters vs. adding a checkpoint, and choose one that incurs lower storage overheads. Once all of the solutions to all of the sub-regions are found, they can be combined into the input. 

\paragraph{Parameter Solving \;}
The parameter (i.e., filters) solving is based on the observation that in a convolution layer, the filter is used \(G^2\) times on varying sub-regions of the input, allowing the creation of \(G^2\) equations representing the use each filter for all \(Y\) filters. To recover all  parameters successfully, the size of the output of a singular filter needs to be greater than the size of a filter, i.e. \(G^2 \geq F^2Z\). If \(G^2 < F^2Z\), padding with dummy input can be used to generate more equations to make the system solvable. 

As the number of channels grows, $F^2Z$ can become much greater than $G^2$, requiring a large number of dummy input to to make the system of equations solvable. As a result, we explore an alternative approach that leads to partial recoverability of parameters. Specifically, since it is extremely rare for a large number of errors in a single layer~\cite{Fiala2011DetectionAC}, we can relax the error recovery capability to recovering up to $G^2$ parameters per filter. To achieve this, we need to be able to identify which group of parameters have errors, as discussed next. Once the erroneous weights are identified, one can create a system equations only representing their effect on the output, reducing the variables in the equation. Allowing the recovery without the additional dummy data. 

\paragraph{Error Detection \;}
In the convolution, each filter produces \(G^2\) output values. Storing just one of the outputs for each filter allows one to monitor if the parameters change. As the input will always be the same, if one of the parameters change, the new output value will differ from the stored value, allowing error detection of a  layer.
\begin{figure}
    \centering
    \includegraphics[scale=.33]{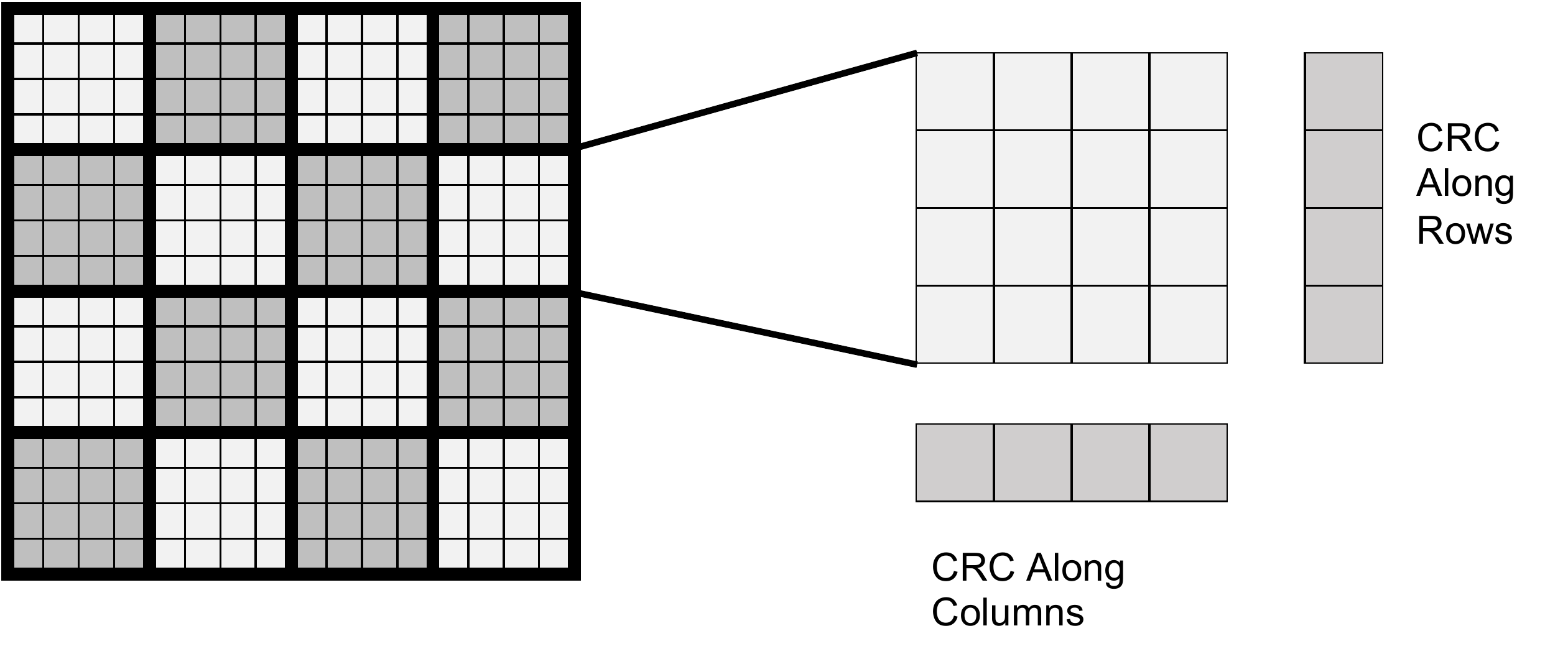}
    \caption{Two dimensional CRC}
    \label{fig:Crc}
\end{figure}

To support parameter solving without adding dummy input and output, it is not sufficient detecting whether a layer has errors in parameters or not. We need to identify the parameters that contain errors. To achieve this, we use a modified version of 2D Error Coding proposed by Kim et al.\cite{Kim2007MultibitET}. In our version, we use cyclic redundancy check (CRC) horizontally and vertically on sets of 4 parameters(Figure~\ref{fig:Crc}), along the last two axis  of the 4D parameter matrix. This is performed \(F^2\) times to fully encode all parameters in the matrix. Encoding along the last two axis of the matrix allows for false positives to be distributed among the filters.

When a layer is identified as erroneous, its CRC codes are recomputed and compared to the stored values.  CRC codes that do not match their stored values are matched up with the CRC codes along the other axis identifying singular weights that are erroneous. This allows for the recovery of only erroneous parameters.  Two Dimensional CRC error detection achieves a low false positive rate as shown in the evaluation. The identified group of erroneous parameters reduce the number of unknowns in the system of equations, allowing recovery. 

\subsection{Pooling Layers}
Pooling layers reduce the dimensionality of the inputted 3d tensor of shape (M, M, Z), where Z is the number of channels. The input is divided into sub-regions of specified size, operating along each channel independently. Each of these sub-regions are used to compute an output based on a singular function. These functions vary with different pooling layers, and many of them use an average or max value function. A pooling layer changes the input in a  non-invertible way. Hence, it requires the addition of a checkpoint that stores the input to the layer. Furthermore, since this layer has no parameters, there is no requirement for a parameter solving function. 

\subsection{Activation Layers}
Convolutional neural networks primarily use ReLu activation layers as they introduce non-linearity into the network and also address the vanishing and exploding gradient problems \cite{Wu2017IntroductionTC, Pedamonti2018ComparisonON}. However, other activation functions can be used throughout the network, both as separate layers and  as parts of other layers. However, all activation layers do have on thing in common: they do not contain any parameters, hence Removing the need for a parameter solving function. As their is such a variety of functions one cannot not say whether an activation layer will be invertible, as it depends on the specific application. But as they do not change the shape of the input as it passes through the layer, during the initialization phase and the error recovery phase all activation functions are treated as linear activation functions. Allowing forward and backward passes through the layer without any changes to the tensor passing through. 

\subsection{Bias Layers}
The bias layer is not technically its own layer. But a part of other layers, such as the convolution and dense layer. But for our work it will be considered its own layer, as it has it own mathematical operation, and its own relationship between its input, output and parameters. 
\begin{equation}
\label{eq:bias}
   Input + Parameters = Output
\end{equation}

The bias layer operates very simply, adding its parameters to the input. This creates a minor shift in values. The bias operation can be represented by equation \ref{eq:bias}. The way that the bias is added can vary slightly based on the layer its connected to; as the bias is a 1D tensor, and the input can vary in dimensionality. For example, in the convolution layer each filters output has a different bias value that is added to all of its outputs. This differs from the dense layer where each element column of the output has its own bias value to be added. 

\paragraph{Backward Pass}
As the layer does simple addition, the subtraction from the parameters from the Output yields the input. Making a backwards pass very fast and efficient.

\paragraph{Parameter Solving}
Parameters solving is also very simple with subtracting the input from output yields the parameters. However, due to the different ways the bias is added based on the layer its attached to, the yielded parameters need to cleaned, removing the duplicate copies, yielding the 1D tensor containing the proper values. 

\paragraph{Error Detection}
Due to the small number of bias parameters we can use a different scheme for error detection compared to other layers. In this layer the sum of all the bias parameters is taken and stored. Therefore if a bias value is changed, the sum would also change detecting an error. There are cases in which two values can change in equal opposite amounts not allowing for error detection. This however is seen to be very unlikely. This scheme saves storage space as in schemes similar to other layers, exploiting parameter reuse, the storage space needed would be equivalent to storing a second copy of the parameters. 

\paragraph{Other Layers}
In a convolutional neural network other layer can and do sometime get used. These layers can include flatten layers, input layers, dropout layers, and padding layers. These layers have different effects on the network and are used for various reasons. In general these layers do not have parameters so they do not need to have a parameter solving function or error detection. For layers that are there for training, and just pass through during prediction such as a dropout layer, they can be essentially ignored. Letting backwards passes pass through them. For layers that adjust the shape, without loosing data such as a flatten or padding layer, on a backwards pass the data will be reshaped to the original form. If data is lost on forward pass, then a checkpoint is stored removing the invertibility requirement. 
\section{Evaluation}
\subsection{Evaluation Method}
Our testing was done on three convolutional neural networks using two different datasets. One network was trained using the MNIST dataset \cite{LeCun2005TheMD}. Two networks a small and a large one were trained using the CIFAR-10 \cite{Krizhevsky2009LearningML} dataset. Details of all three networks are shown in tables \ref{tab:MNIST}, \ref{tab:cifar}  and \ref{tab:cifar_B}. 
\begin{scriptsize}
    \begin{table}[htbp]
        \centering
        \caption{MNIST network }
        \label{tab:MNIST}
        \begin{tabular}{|l|l|l|}
            \hline
            \textbf{Layer} & \textbf{Output Shape} & \textbf{Trainable}\\
            \hline
            \hline
            Conv. 2D    & (26,26,32)    & 320   \\
            \hline
            Conv. 2D    & (24,24,32)    & 9,248  \\
            \hline
            Max Pooling & (12,12,32)    & 0     \\
            \hline
            Conv. 2D    & (10,10,64)    & 18,496  \\
            \hline
            Dense       & (256)         & 1,638,656 \\
            \hline
            Dense       & (10)          & 2,570 \\
            \hline
        \end{tabular}
    \end{table}
\end{scriptsize}
\begin{scriptsize}
    \begin{table}[htbp]
        \caption{CIFAR-10 small network }
        \label{tab:cifar}
        \centering
        \begin{tabular}{|l|l|l|}
            \hline
            \textbf{Layer} & \textbf{Output Shape} & \textbf{Trainable}\\
            \hline
            \hline
            Conv. 2D    & (32,32,32)    & 896   \\
            \hline
            Conv. 2D    & (32,32,32)     & 9,248  \\
            \hline
            Max Pooling & (16,16,32)    & 0     \\
            \hline
            Conv. 2D    & (16,16,64)    & 18,496  \\
            \hline
            Conv. 2D    & (16,16,64)    & 36,928  \\
            \hline
            Max Pooling & (8,8,64)    & 0     \\
            \hline
            Conv. 2D    & (8,8,128)    & 73,856  \\
            \hline
            Conv. 2D    & (8,8,128)    & 147,584  \\
            \hline
            Conv. 2D    & (8,8,128)    & 147,584 \\
            \hline
            Max Pooling & (4,4,128)    & 0     \\
            \hline
            Dense       & (128)         & 262,272\\
            \hline
            Dense       & (10)          & 1,290 \\
            \hline
        \end{tabular}
    \end{table}
\end{scriptsize}
\begin{scriptsize}
    \begin{table}[htbp]
        \caption{CIFAR-10 large network }
        \label{tab:cifar_B}
        \centering
        \begin{tabular}{|l|l|l|}
            \hline
            \textbf{Layer} & \textbf{Output Shape} & \textbf{Trainable}\\
            \hline
            \hline
            Conv. 2D    & (32,32,96)    & 7,296   \\
            \hline
            Max Pooling & (16,16,96)    & 0     \\
            \hline
            Conv. 2D    & (16,16,96)     & 230,496  \\
            \hline
            Max Pooling & (8,8,96)    & 0     \\
            \hline
            Conv. 2D    & (8,8,80)    & 192,080  \\
            \hline
            Conv. 2D    & (8,8,64)    & 128,064  \\
            \hline
            Conv. 2D    & (8,8,64)    & 102,464  \\
            \hline
            Conv. 2D    & (8,8,96)    & 153,696  \\
            \hline
            Dense       & (256)         & 1,573,120\\
            \hline
            Dense       & (10)          & 2,570 \\
            \hline
        \end{tabular}
    \end{table}
\end{scriptsize}

We perform three types of experiments: (1) inject bit errors a probability of \(p\) (i.e. Raw Bit Error Rates (RBER)), (2) inject whole-weight errors with a probability of \(q\), and (3) corrupt entire layers. Experiment (1) is to the effectiveness of MILR in comparison to ECC if no memory encryption is used, which gives a rough idea of MILR capability in traditional random bit error settings. Experiments (2) and (3) are most relevant in plaintext space error correction (PSEC) that is the goal of MILR. In plaintext space, ECC is unable to recover from such errors, while MILR can in many cases. The injection of bit errors is done by flipping each bit with a probability \(p\) (error rate), regardless of bit position and role (each 32-bit float parameter has sign, magnitude and mantissa).  Whole-weights are injected by flipping every bit in a weight with a probability of \(q\). Entire layers are corrupted by replacing the entire layers parameters with new random parameters. These experiments attempt to simulate plaintext space errors and plaintext-level security attacks. The random bit flipping simulates soft memory errors, and to a more limited extent security attack such as it Flip Attacks~\cite{Rakin2019BitFlipAC} or fault injection attacks \cite{Liu2017FaultIA}, with the whole-weight and layer corruption focusing on more aggressive security attacks. 

MILR was compared to SECDED (single-bit error correction and double-bit error detection) ECC protecting each word. This (39,32) code requires 7 additional ECC bits for each 32-bit word that coincides with a single parameter, allowing error recovery for any parameter if a single bit of it is corrupted. In the case of more than 1 bit error no correction occurs and interrupts is not raised. 

\textbf{Implementation}
MILR was implemented as a library that could be used with TensorFlow~\cite{tensorflow2015-whitepaper} taking a Tensorflow model as input. The model is initially processed to prepare for error detection and recovery. Periodically, MILR's error detection function is called, and if errors are detected, the error recovery function is also called. MILR can recover any number of errors in a single layer between a pair of checkpoints for dense and convolution layers, or \(G^2\) parameters for convolution layers using partial recoverability. However, if the RBER is very high, there may be more than one erroneous layers between a pair of checkpoints. In this case, full self-healing cannot be guaranteed. However, error recovery is invoked regardless, and applied to the erroneous layers in sequential order. 

\textbf{Limitations}
As MILR was implemented as an external library and not as a part of TensorFlow execution pipeline limiting MILR's performance. MILR does take advantage of TensorFlow's function calls where possible, but further performance optimization may be possible. MILR error recovery relies on algebraic relationship of input, parameter, and output, hence it is affected by floating point arithmetic rounding in binary representations, e.g. algebraically \((a+b)+c = a+(b+c)\) but with binary floating point representation and rounding,  \((a+b)+c \approx a+(b+c)\). This is made worse in large computation for solving large systems of equations.

\subsection{MNIST Network}
The MNIST handwritten numbers database \cite{LeCun2005TheMD} is a commonly used database in machine learning, with 60,000 28 x 28 black and white training images and 10,000 images for testing; which are classified into 10 categories. The network was built according to Table~\ref{tab:MNIST}, with valid padding convolution layers, and a bias and ReLu activation layer after each dense and convolution layer. The network was trained for 5 epochs with a batch size of 128, to an accuracy of \(99.2\%\).
\begin{figure*}[htbp]
    \scriptsize
    \centering
    \subfloat[No recovery]{
        \begin{minipage}{0.235\textwidth}
            \centering
            \includegraphics[width=\textwidth]{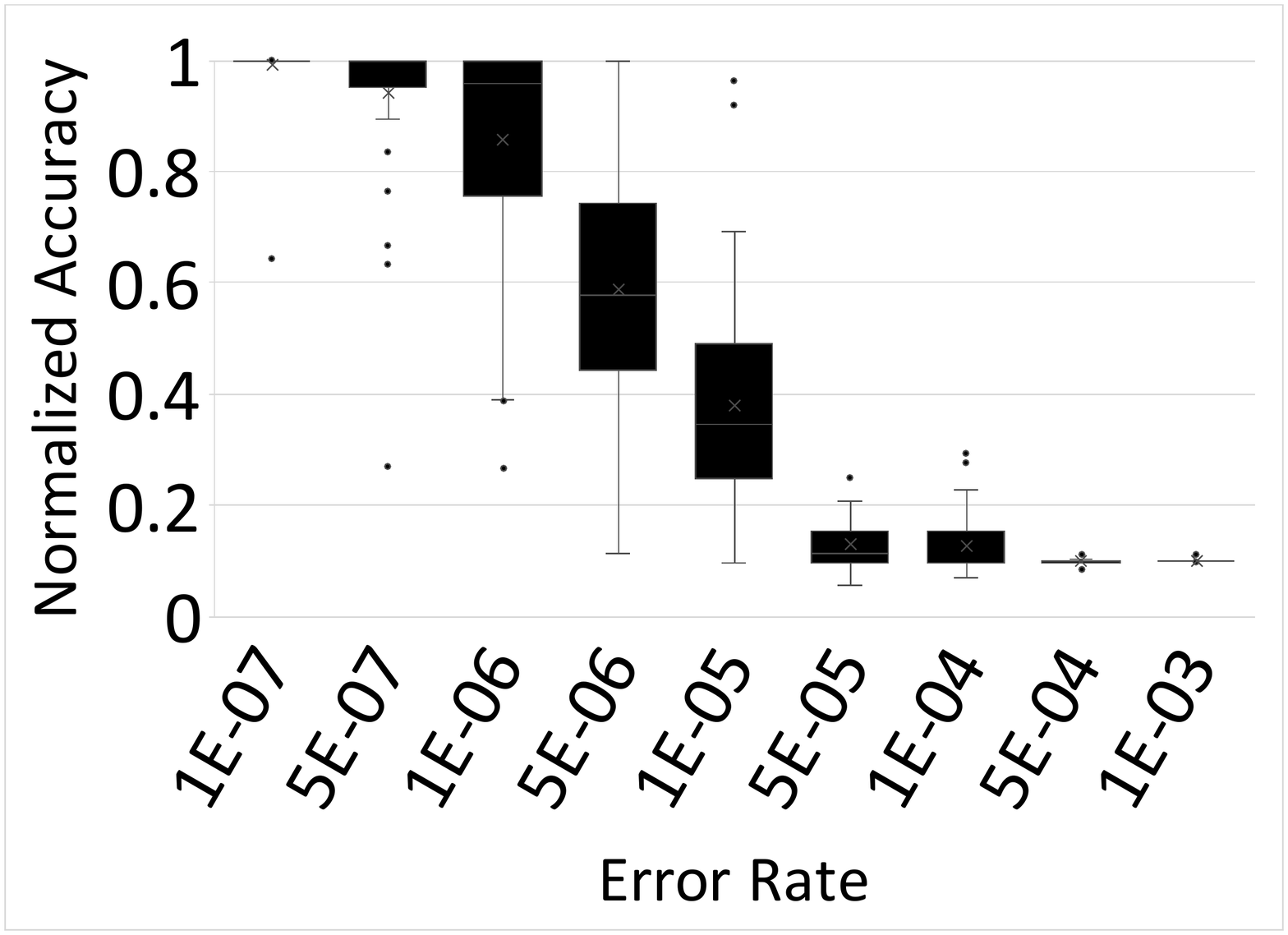}
            \label{fig:MNIST_error}
        \end{minipage}
    }
    \subfloat[ECC]{
        \begin{minipage}{0.235\textwidth}
            \centering
            \includegraphics[width=\textwidth]{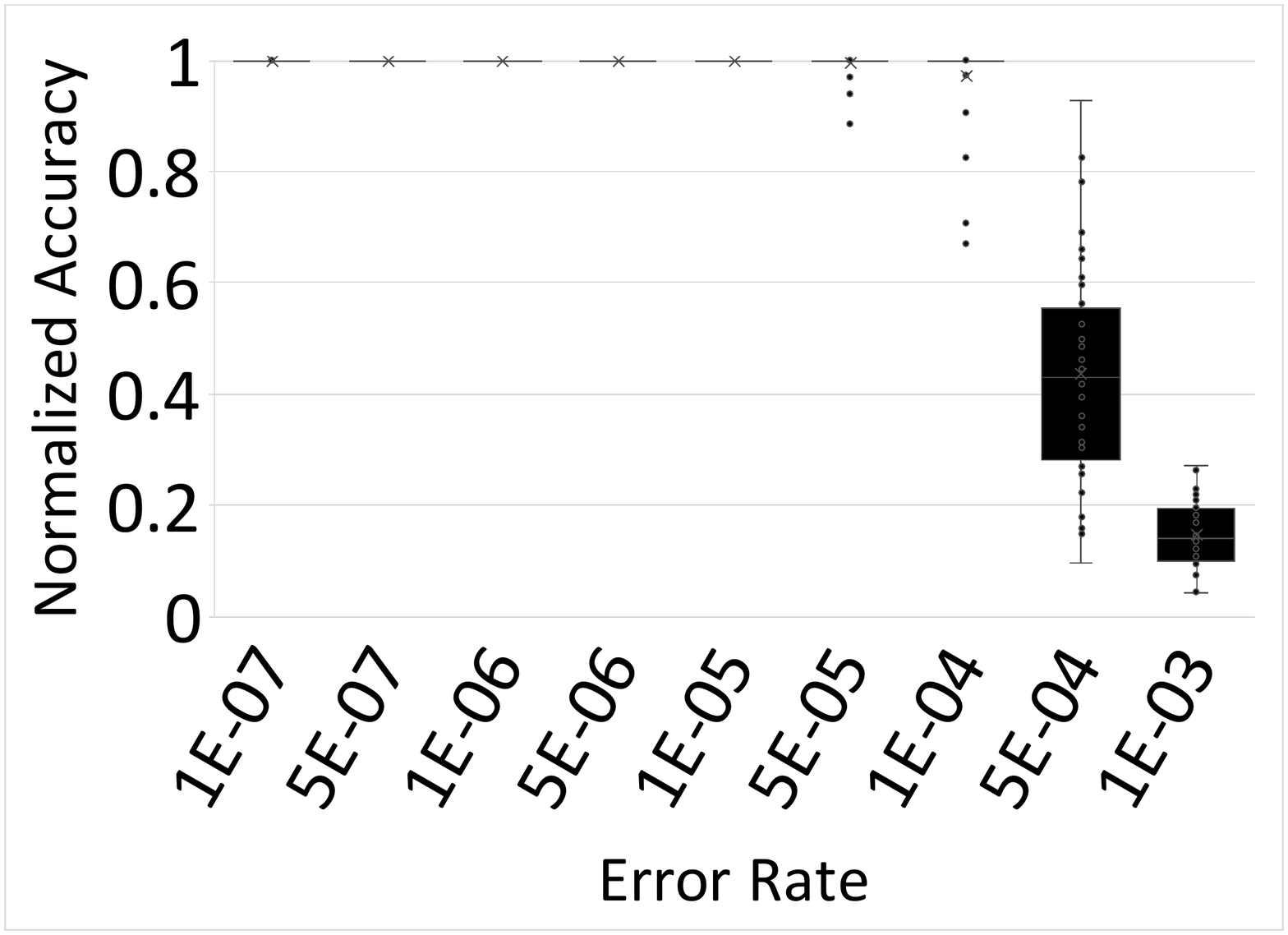}
            \label{fig:MNSIT_ecc}
        \end{minipage}
    } 
    \subfloat[MILR]{
        \begin{minipage}{0.235\textwidth}
            \centering
            \includegraphics[width=\textwidth]{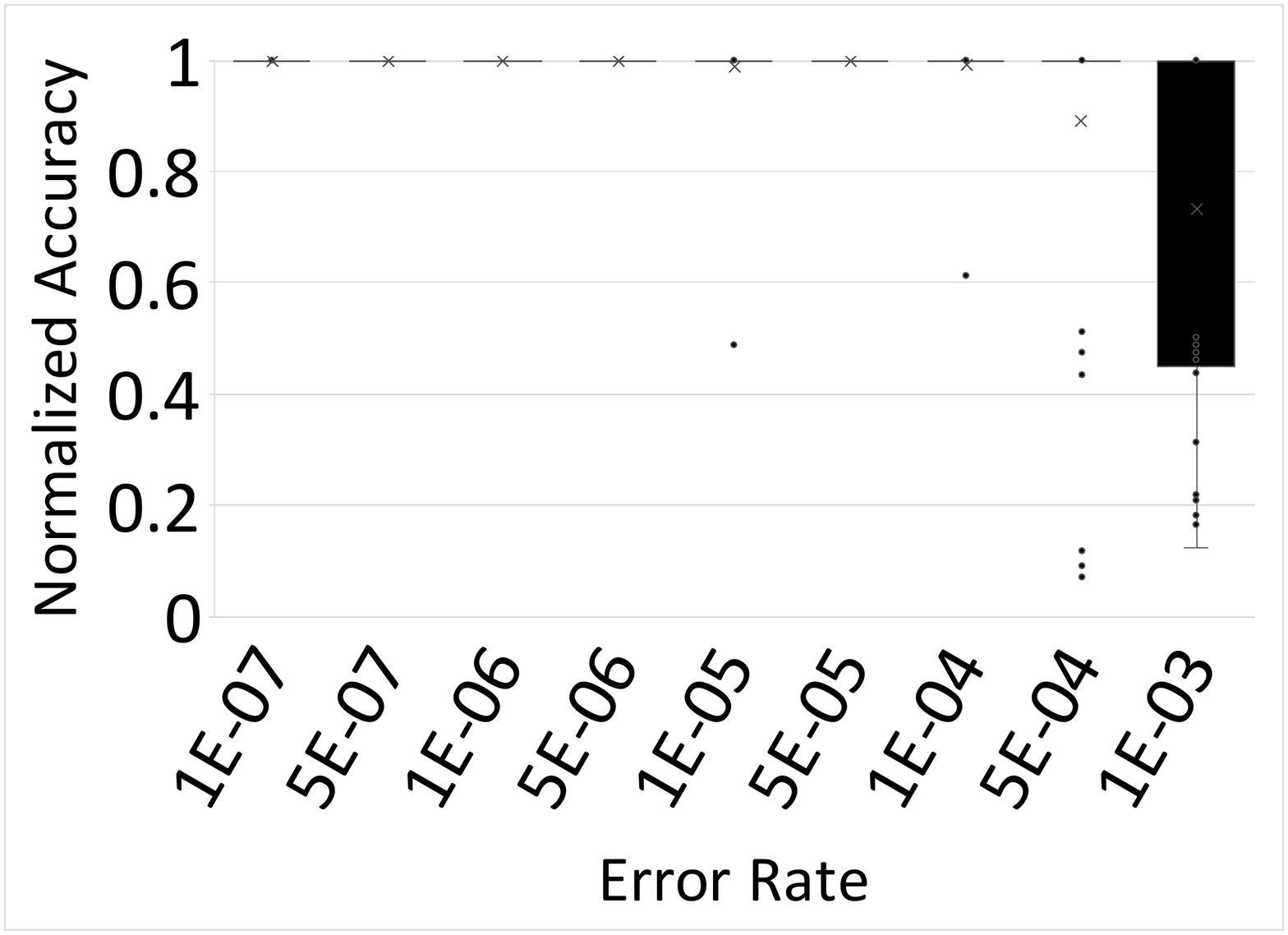}
            \label{fig:MNIST_recov}
        \end{minipage}
    }
    \subfloat[ECC + MILR]{
        \begin{minipage}{0.235\textwidth}
            \centering
            \includegraphics[width=\textwidth]{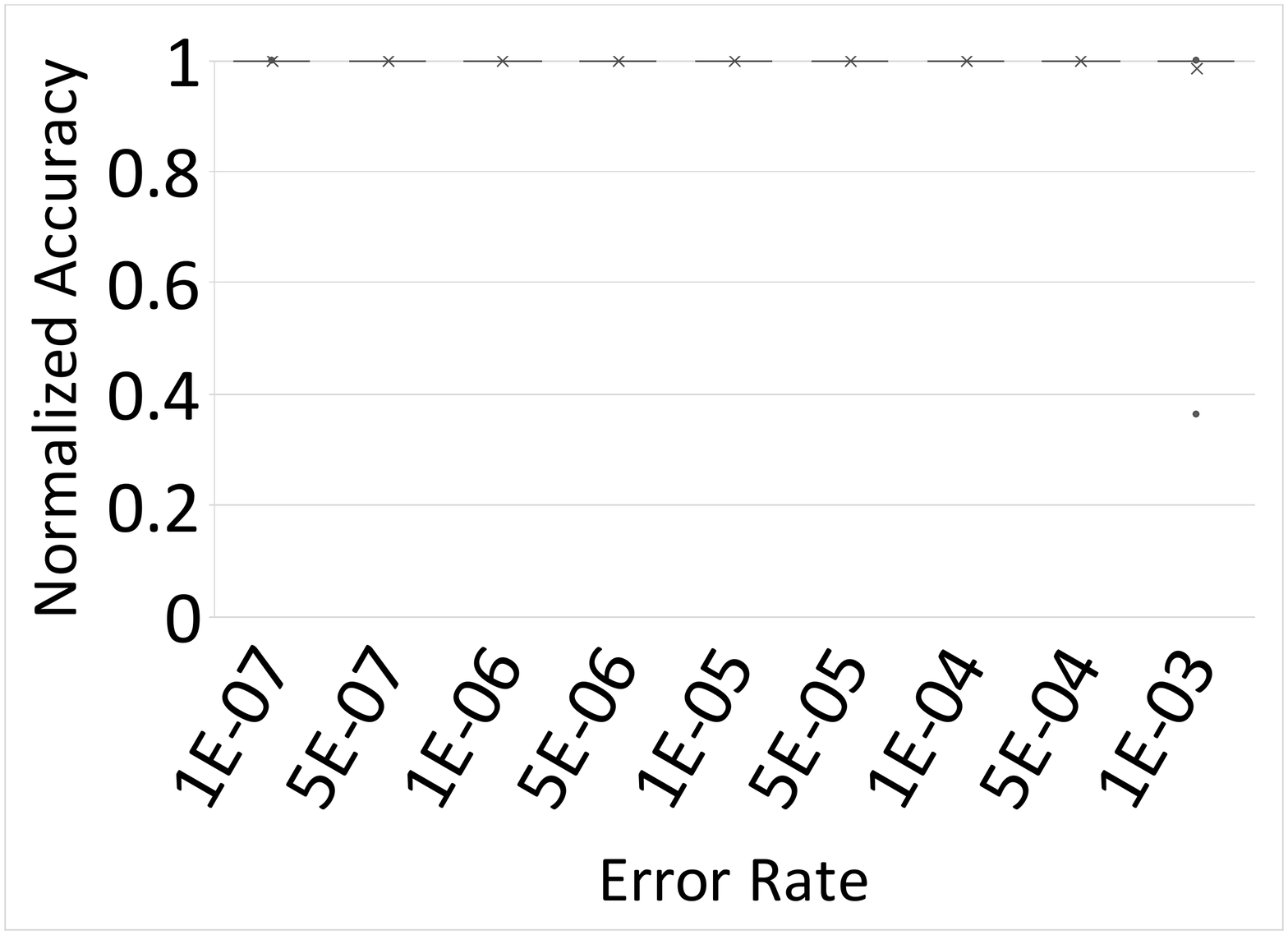}
            \label{fig:mnist_eccMilr}
        \end{minipage}
    }
    \caption{MNIST network normalized accuracy after recovery from varying RBER}
    \label{fig:MNIST-ACC}
\end{figure*}

Figure~\ref{fig:MNIST-ACC} shows box plots of 40 runs with varying RBER with normalized average accuracy (i.e. 100\% means the same accuracy as the error-free version). Figure~\ref{fig:MNIST_error} shows the raw un-recovered effect of the RBER, with figures \ref{fig:MNSIT_ecc}, and \ref{fig:MNIST_recov}, and \ref{fig:mnist_eccMilr} 
showing the accuracy after error is detected and recovered using ECC, MILR and ECC + MILR.Each plot is centered on the median values, with the box covering the \(25^{th}\) and \(75^{th}\) percentile (i.e. {\em interquartile} range). The whiskers extended \(1.5\times\) the interquartile range from the top and bottom of the box, up to the max/min value. Outliers are marked by dots on the graph.

The MNIST network (Figure~\ref{fig:MNIST_error}) has a little bit of built-in robustness to errors, keeping accuracy high (99.1\%) for 1E-07 RBER. However, in some cases with 5 flipped bits, the normalized accuracy drops to \(64.5\%\). This is because not all bit positions are equal in their impact on accuracy, most significant bit (MSB) has a larger impact on accuracy than least significant bit (LSB)\cite{Qin2017RobustnessON}. ECC increases the robustness of the network (Figure~\ref{fig:MNSIT_ecc}), but as RBER increases such that multi-bit errors occur (after \(1E-05\)), ECC's performance starts to drop.

MILR is able to increase the robustness of the network over the no recovery and ECC as it provides \(99\%\) of accuracy through \( 1E-04\), as shown in Figure~\ref{fig:MNIST_recov}, with an outlier at \(1E-05\). This is due to MILR being able to recover from both single and multi bit errors. MILR has some outliers with lower accuracies after \(1E-04\), as the frequency of errors affecting multiple layers between a pair of checkpoints increases. This causes the input/output pairs being used to recover the parameters to be erroneous from having to pass through erroneous layers to get the destination layer, diminishing the accuracy of the recovery.

Investigating the outliers, they are caused by either too many erroneous layers or some errors are not detected. For the former problem, MILR can use more checkpoints, or alternatively, utilize a combination of ECC and MILR. When combined (Figure~\ref{fig:mnist_eccMilr}), ECC addresses most single bit errors before reaching MILR, leaving MILR to deal with multi-bit errors. The removal of the majority of the single bit errors helps prevent multiple erroneous layers between a pair of checkpoints. 

Another cause of outliers with lower accuracy is the error detection limitation. With MILR, before recovery is initiated, error detection phase must identify what layers need to be recovered. Our detection scheme for MILR is a lightweight detection scheme that requires the errors to be significant enough to detect. This does mean that not all errors will be detected; they are only detected when they have a meaningful impact on the output of the layer. For MNIST, in \(78.6\%\) of the tests, all erroneous layers were detected. In the remaining 21.3\% of cases where not all erroneous layer were detected MILR still restores the accuracy to \(99.9\%\) of the original accuracy.  

\begin{figure}[htbp]
    \scriptsize
    \centering
    \subfloat[No recovery]{
        \begin{minipage}{0.235\textwidth}
            \centering
            \includegraphics[width=\textwidth]{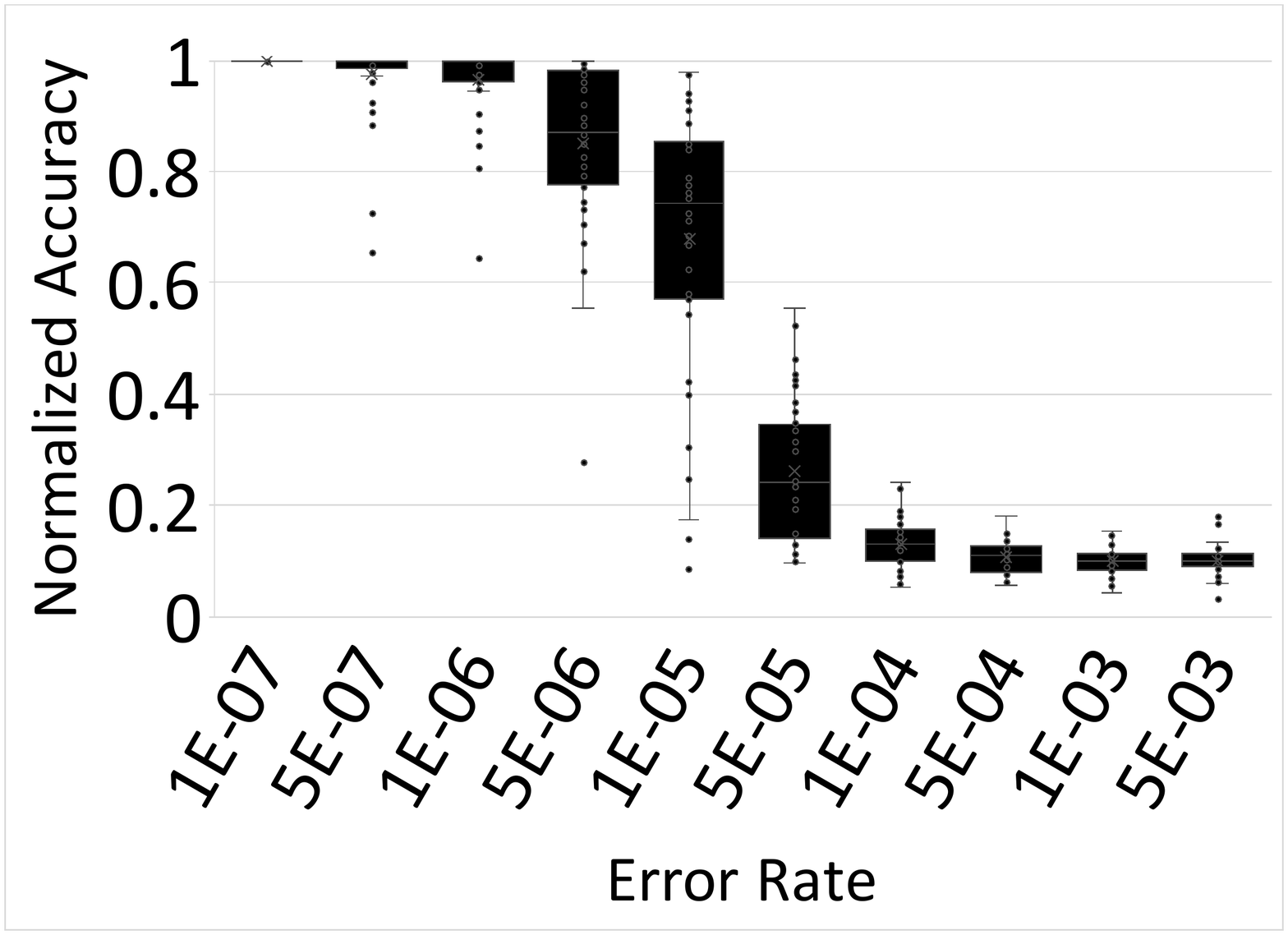}
            \label{fig:MNIST_error_whole}
        \end{minipage}
    }
    \subfloat[MILR]{
        \begin{minipage}{0.235\textwidth}
            \centering
            \includegraphics[width=\textwidth]{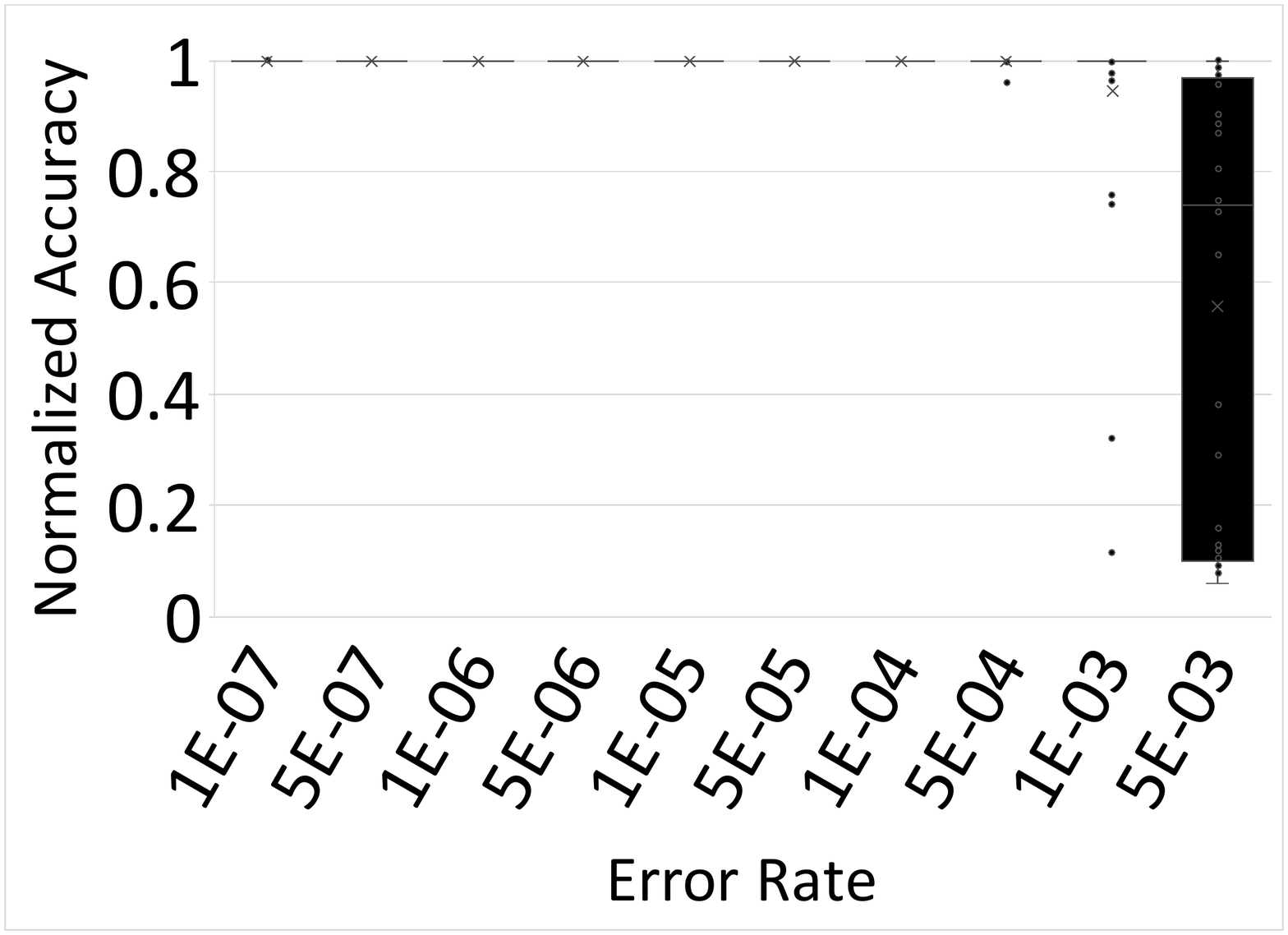}
            \label{fig:MNIST_recov_whole}
        \end{minipage}
    }
    \caption{MNIST network normalized accuracy after recovery from whole-weight errors}
    \label{fig:MNIST-ACC-whole}
\end{figure}
The MNIST network was also tested with whole-weight errors with a probability of \(q\), where every bit in a word was flipped. ECC and ECC + MILR were not tested with this scheme as ECC can only correct 1 bit errors and all errors injected would be 32 bit errors. The network still had some intrinsic robustness with having at least \(97.3\%\) of accuracy through an error rate of \(5E-07\). When MILR is applied it is able to recover the network to \(99.9\%\) accuracy through \(5e-04\). After this point, multiple erroneous layers between checkpoint  pairs start to appear more frequently starting to affect the recovery accuracy.

To test a scenario when a whole layer is erroneous each layer individually has all of its parameters replaced by a random values, where none of the values were the same as the original value. Then MILR attempted to solve the layers back to its original state. For the convolution layers using the  partial recoverability by design they can not recovery the full layer. In those scenarios, they followed their standard recovery process but when they attempt to solve their system of equation they have more variables than equations. To address this they attempt to find a least-square solution. This provides a solution to the linear equation as close as possible to the actual solution. The less under-defined the problem the close the solution should be, however the precision will vary. If the system of equation is to underdefined the system is not solvable even by finding the least-square solution. In these cases no recovery is possible, however the probability of these cases in real world scenarios are slim.  Also in this test, the bias of a layer is considered as separate layer and is treated as such. 
\begin{scriptsize}
    \begin{table}[htbp]
        \centering
        \caption{MNIST network whole layer error accuracy}
        \label{tab:MNIST_acc_wholelayer}
        \begin{threeparttable}
            \begin{tabular}{|l|l|l|}
                \hline
                \textbf{Layer} & \textbf{None} & \textbf{MILR}\\
                \hline
                \hline
                Conv. & 46.9\% & 100.0\%\\
                \hline
                Conv. Bias & 75.7\% &100.0\%\\
                \hline
                Conv. 1 & 34.9\% & N/A \tnote{*}\\
                \hline
                Conv. 1 Bias &  81.7\% & 100.0\%\\
                \hline
                Conv. 2 &  23.1\% & N/A \tnote{*}\\
                \hline
                Conv 2 Bias &    77.4\% & 	100.0\%\\
                \hline
                Dense &    10.2\% & 100.0\%\\
                \hline
                Dense Bias &    100.0\% & 100.0\%\\
                \hline
                Dense 1 &    9.9\% & 100.0\%\\
                \hline
                Dense 1 Bias &    100.0\% &100.0\%\\
                \hline
            \end{tabular}
            \begin{tablenotes}
            \item[*]Convolution partial recoverable
            \end{tablenotes}
        \end{threeparttable}
    \end{table}
\end{scriptsize}

Whole layer errors showed that each layer is important to the network serving a purpose. The main function of the layer is of key performance, with the bias layer serving less but still playing a significant role. The bias layers of the dense layers had the least affect on the network compared to other layers. For recovery MILR was able to recovery all complete recoverable layers complete. For the convolutional layers using partial recoverability they were to underdefined to be recovered. 
\begin{scriptsize}
    \begin{table}[htbp]
        \centering
        \caption{MNIST network storage overhead }
        \label{tab:MNIST_storage}
        \begin{tabular}{|l|l|l|l|}
            \hline
            \textbf{Backup Weights} & \textbf{ECC} & \textbf{MILR}& \textbf{ECC \& MILR}  \\
            \hline
            \hline
            6.68 MB & 1.46 MB & 6.81 MB & 8.27 MB\\ 
            \hline
        \end{tabular}
    \end{table}
\end{scriptsize}

 MILR's storage of additional data needed can vary from network to network, as it varies based on the networks structure. Hyperparameters such as layer order, layer type and layer configuration (filter size, filter count, etc.) can effect the overhead of MILR. An optimized network for MILR can reduce the overhead compared to a non-optimized network, but MILR is able to be applied to any CNN and work as expected. MILR requires to store an additional 6.81 MB of data for MNIST network error detection and recovery. ECC adds 1.46 MB but it has limited error recoverability as well as the overheads are incurred at DRAM working memory. In contrast, MILR storage overheads can be placed in SSD, HDD, or persistent memory, which are orders of magnitude denser and cheaper than DRAM. Keeping a backup copy of the network allows for redundancy but requires just as much storage overhead while not being able to detect errors. 
 
\subsection{CIFAR-10 Small Network}
The Cifar-10 dataset \cite{Krizhevsky2009LearningML} is a color image database with 60,000 \(32\times 32\)  images, that can be categorized into 10 categories with 6,000 examples of each in the dataset. The dataset is partitioned into 50,000 images for training and 10,000 for testing. The small network was built according to Table~\ref{tab:cifar} with same padding convolution layers; and bias and ReLu activation layer after each dense and convolution layer. The architecture of the network was inspired by the VGG network \cite{Simonyan2015VeryDC}, but minus the last few layers and shallower filter depth as we not using the Imagenet database \cite{ILSVRC15}. The CIFAR-10 network was trained to \(84.8\%\) accuracy, over 150 epochs with a batch size of 128.
\begin{figure*}[htbp]
\scriptsize
\centering
    \subfloat[No recovery]{
        \begin{minipage}{0.235\textwidth}
            \centering
            \includegraphics[width=\textwidth]{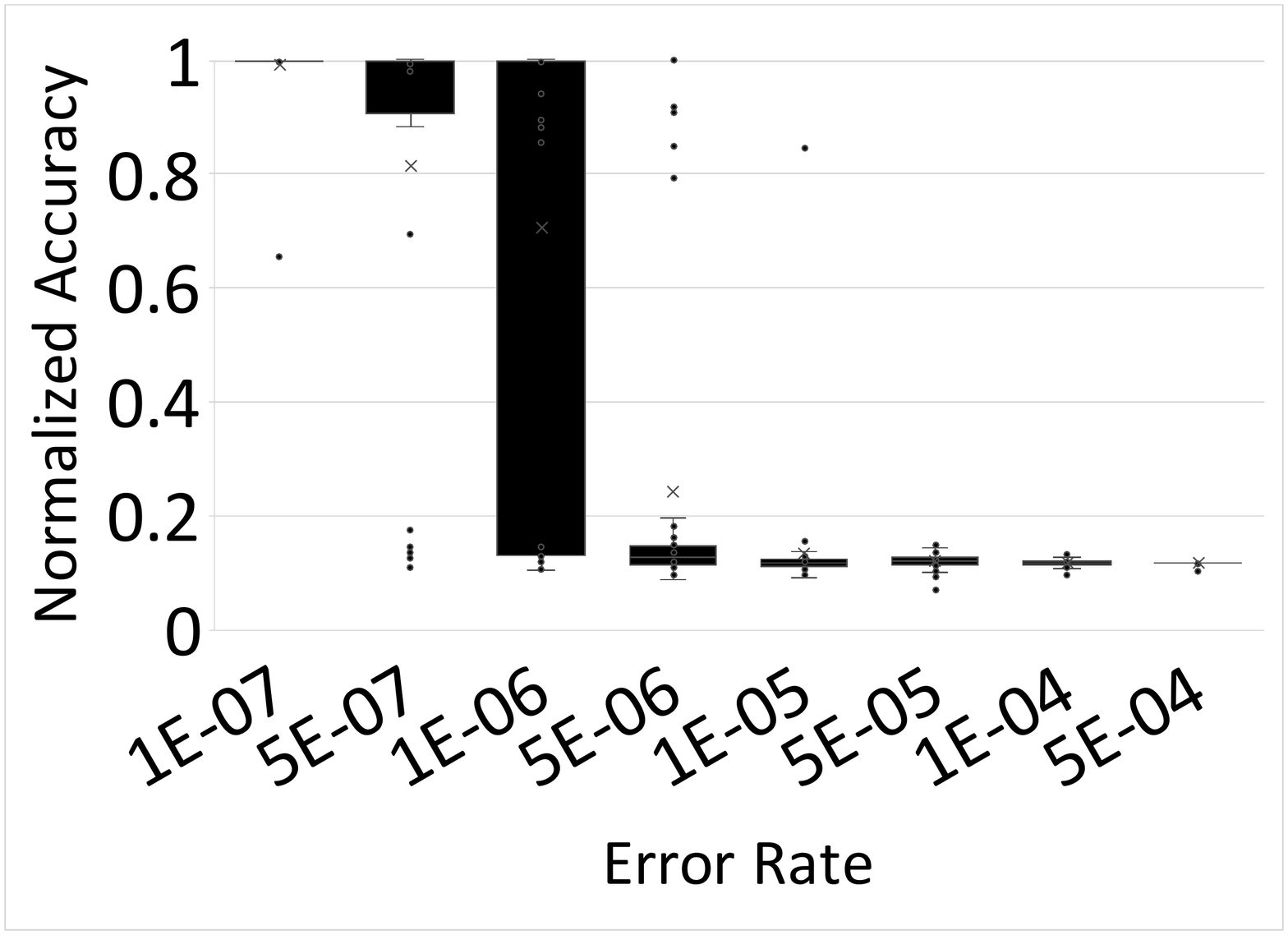}
            \label{fig:cifar_error}
        \end{minipage}
    }
    \subfloat[ECC]{
        \begin{minipage}{0.235\textwidth}
            \centering
            \includegraphics[width=\textwidth]{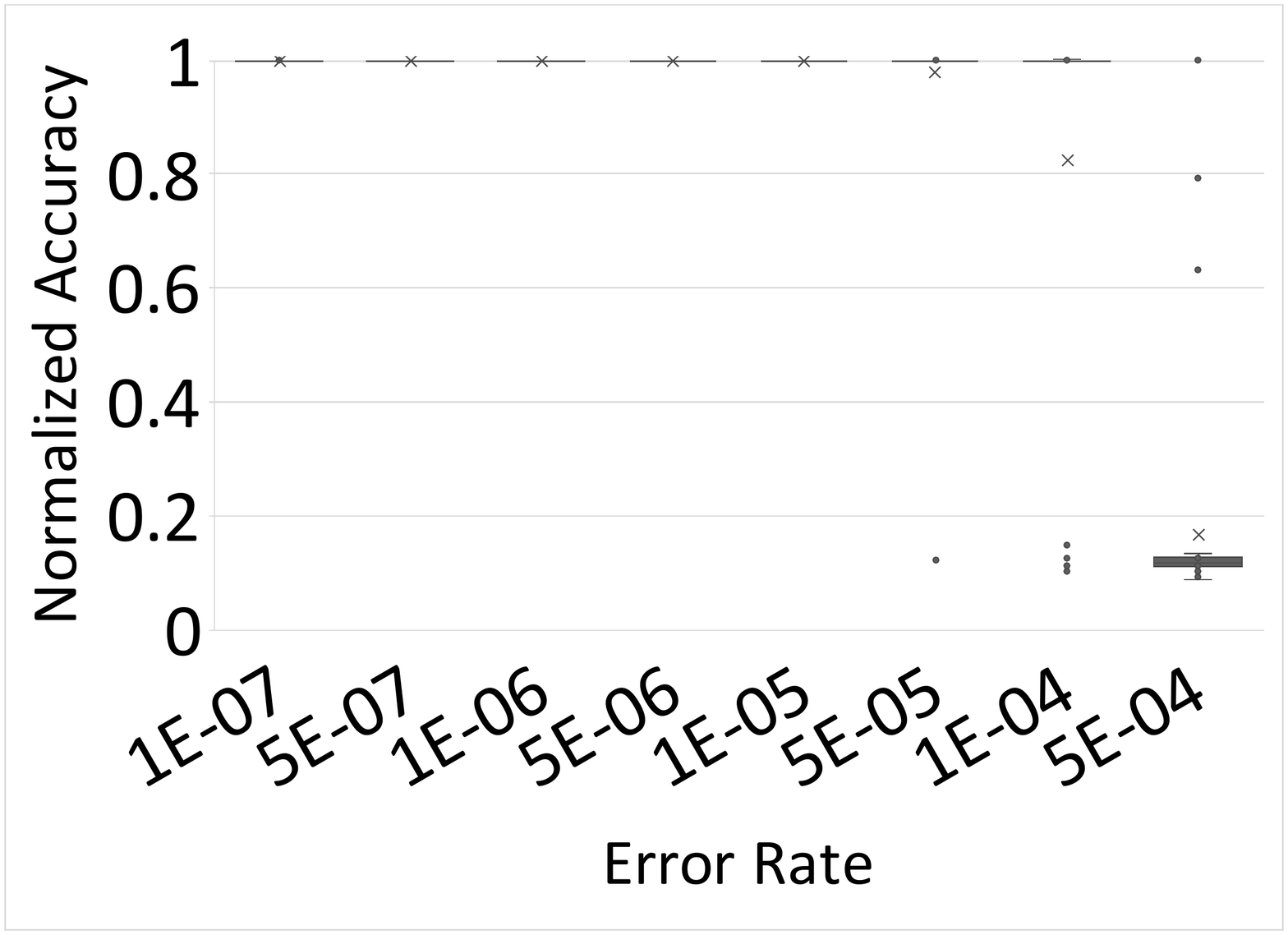}
            \label{fig:cifar_ecc}
        \end{minipage}
    } 
    \subfloat[MILR]{
        \begin{minipage}{0.235\textwidth}
            \centering
            \includegraphics[width=\textwidth]{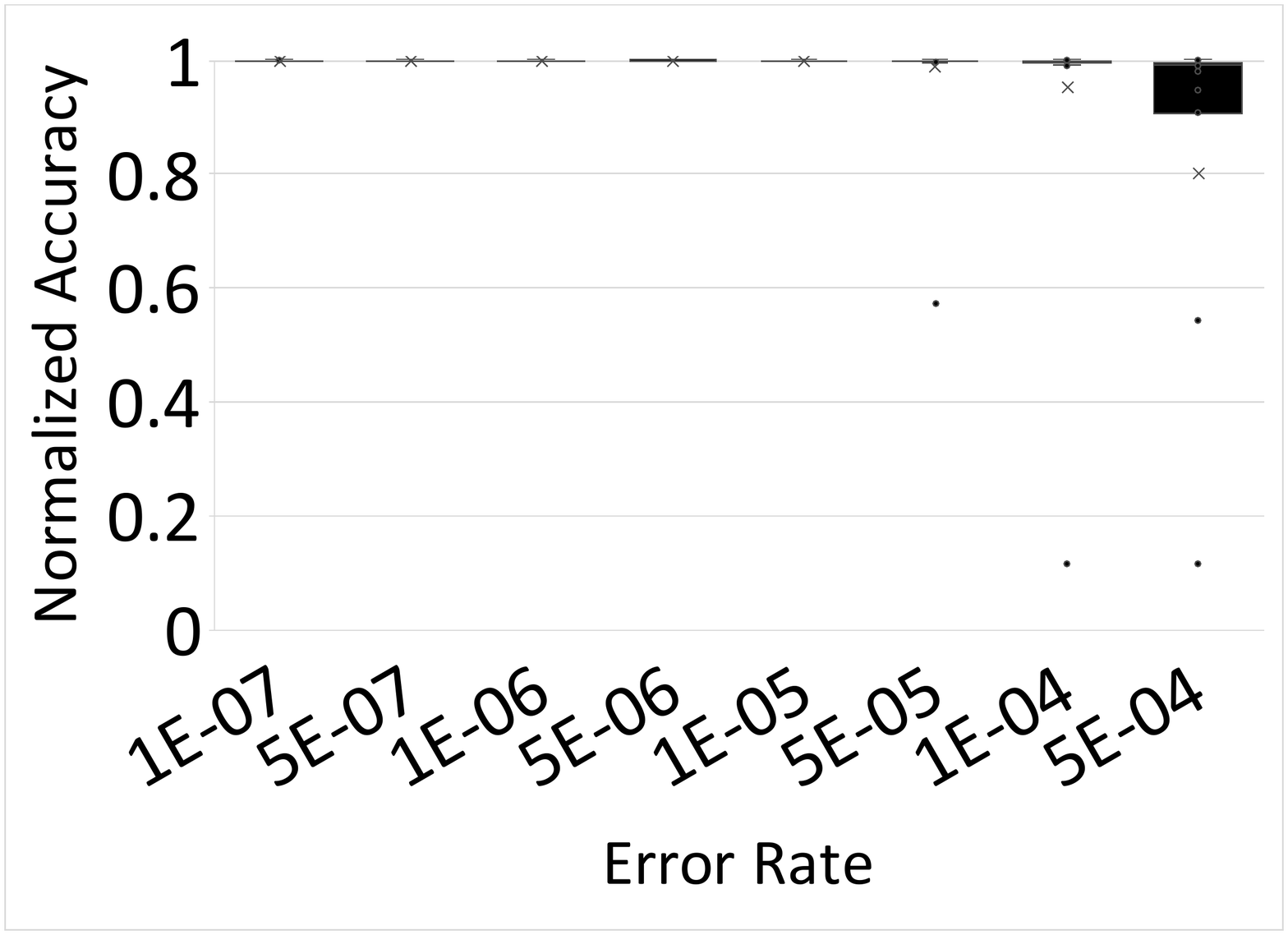}
            \label{fig:cifar_recov}
        \end{minipage}
    }
    \subfloat[ECC + MILR]{
        \begin{minipage}{0.235\textwidth}
            \centering
            \includegraphics[width=\textwidth]{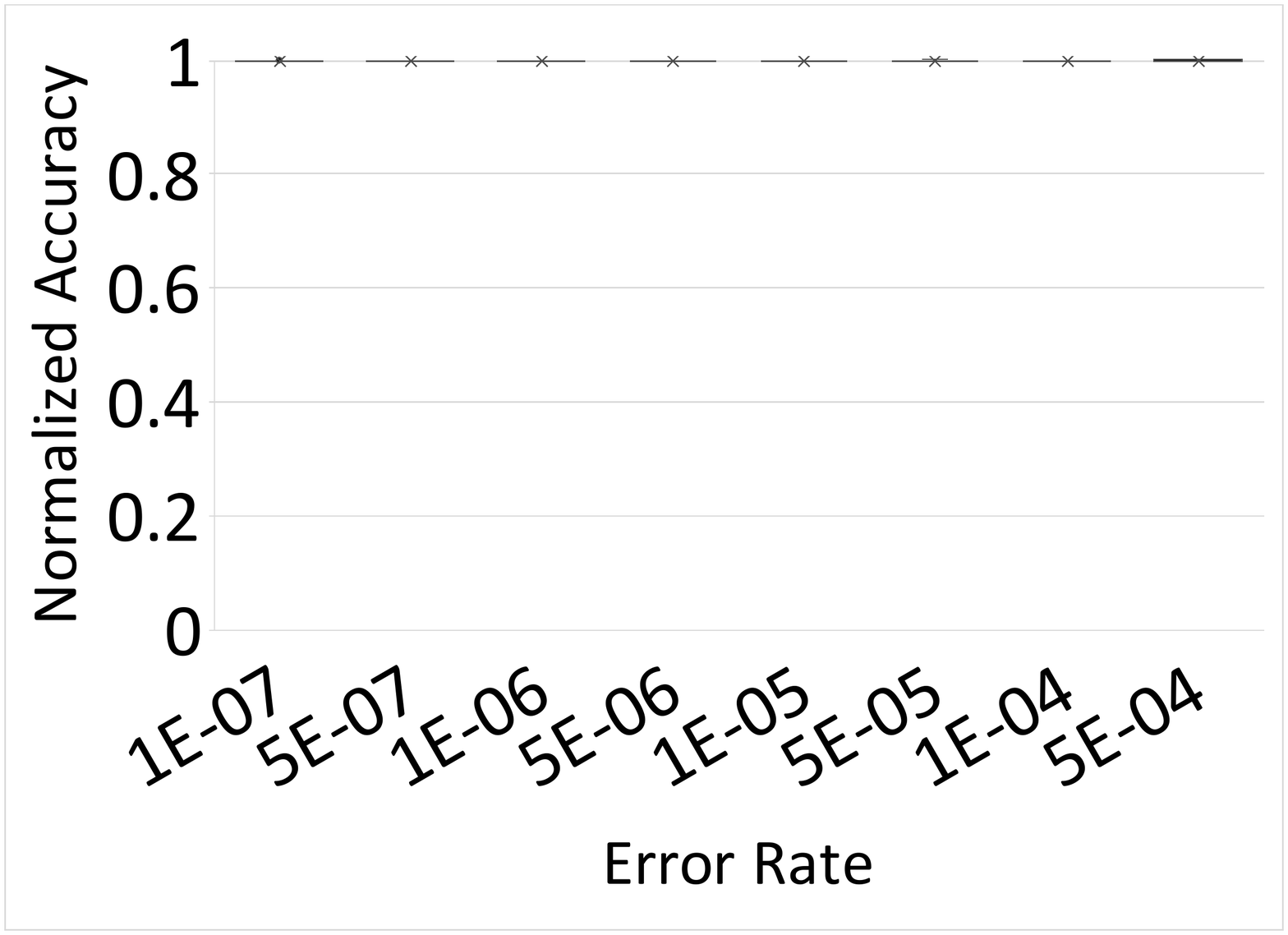}
            \label{fig:cifar_eccMilr}
        \end{minipage}
    }
    \caption{CIFAR-10 small network normalized accuracy after recovery from varying RBER}
    \label{fig:CIFAR-ACC}
\end{figure*}

The CIFAR-10 network does have some intrinsic robustness, shown in Figure~\ref{fig:CIFAR-ACC}, achieving \(99.1\%\) of  accuracy with an error rate of \(1E-07\). ECC maintains \(100.0\%\) of accuracy to  \(1E-05\) as it corrects all single bit errors. But, as with the MNIST network, its accuracy drops as multi-bit errors start to appear. MILR was able to achieve the same performance to \(1E-05\) however it also dropped after this point as with ECC. That being said it dropped less drastically and more gradually as the error increased maintaining \(80.2\%\) accuracy at \(5E-04\) while ECC had only \(16.9\%\) of accuracy. The combination of ECC + MILR was able to recover back to \(100.0\%\) of accuracy through the test range.

For the CIFAR-10 network MILR detected all erroneous layers in \(64.7\%\) of the test. With  \(99.1\%\) of these test being restore to \(> 99.3\%\) of original accuracy.
\begin{figure}[htbp]
    \scriptsize
    \centering
    \subfloat[No recovery]{
        \begin{minipage}{0.235\textwidth}
            \centering
            \includegraphics[width=\textwidth]{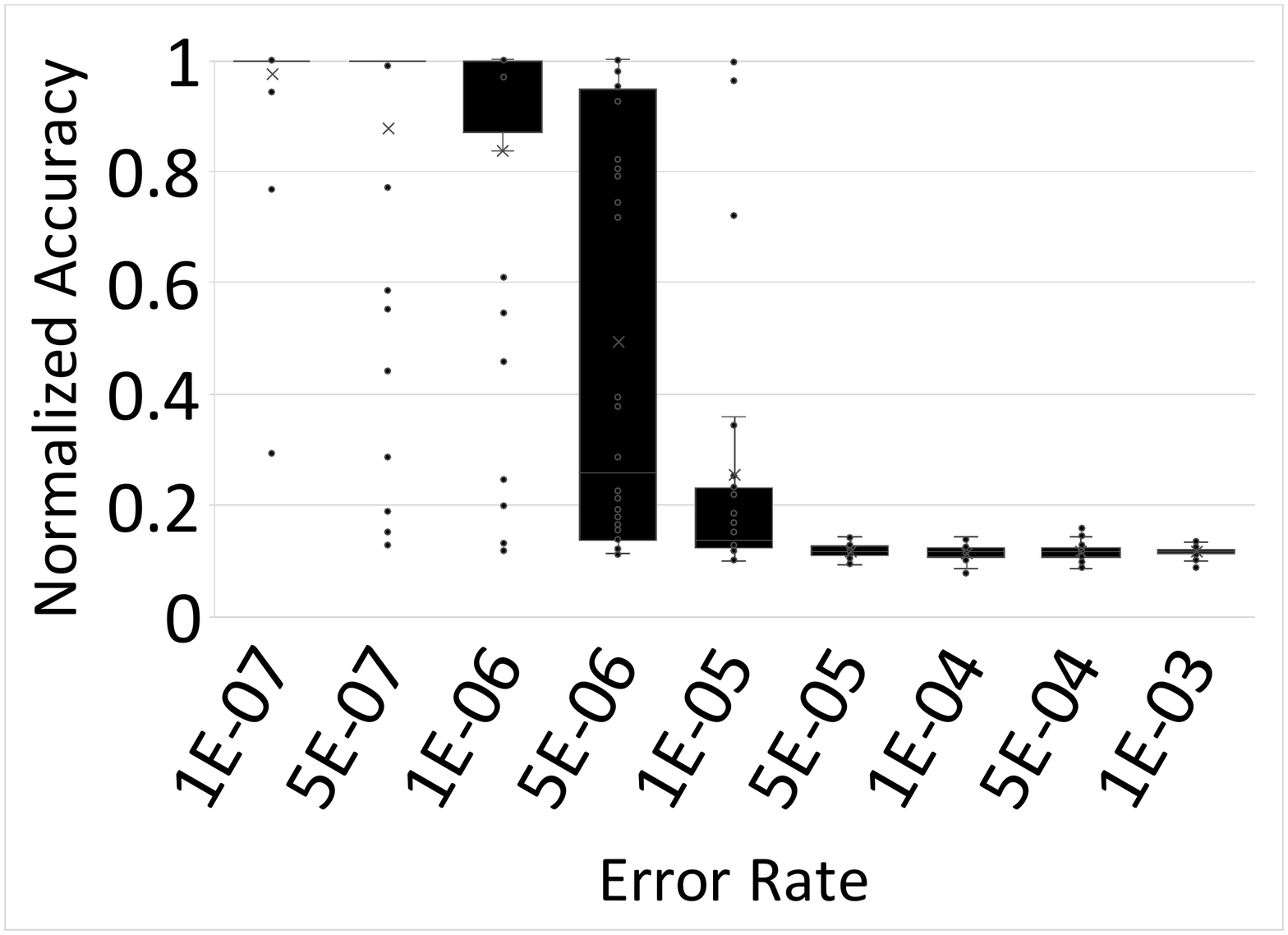}
            \label{fig:CIFAR_error_whole}
        \end{minipage}
    }
    \subfloat[MILR]{
        \begin{minipage}{0.235\textwidth}
            \centering
            \includegraphics[width=\textwidth]{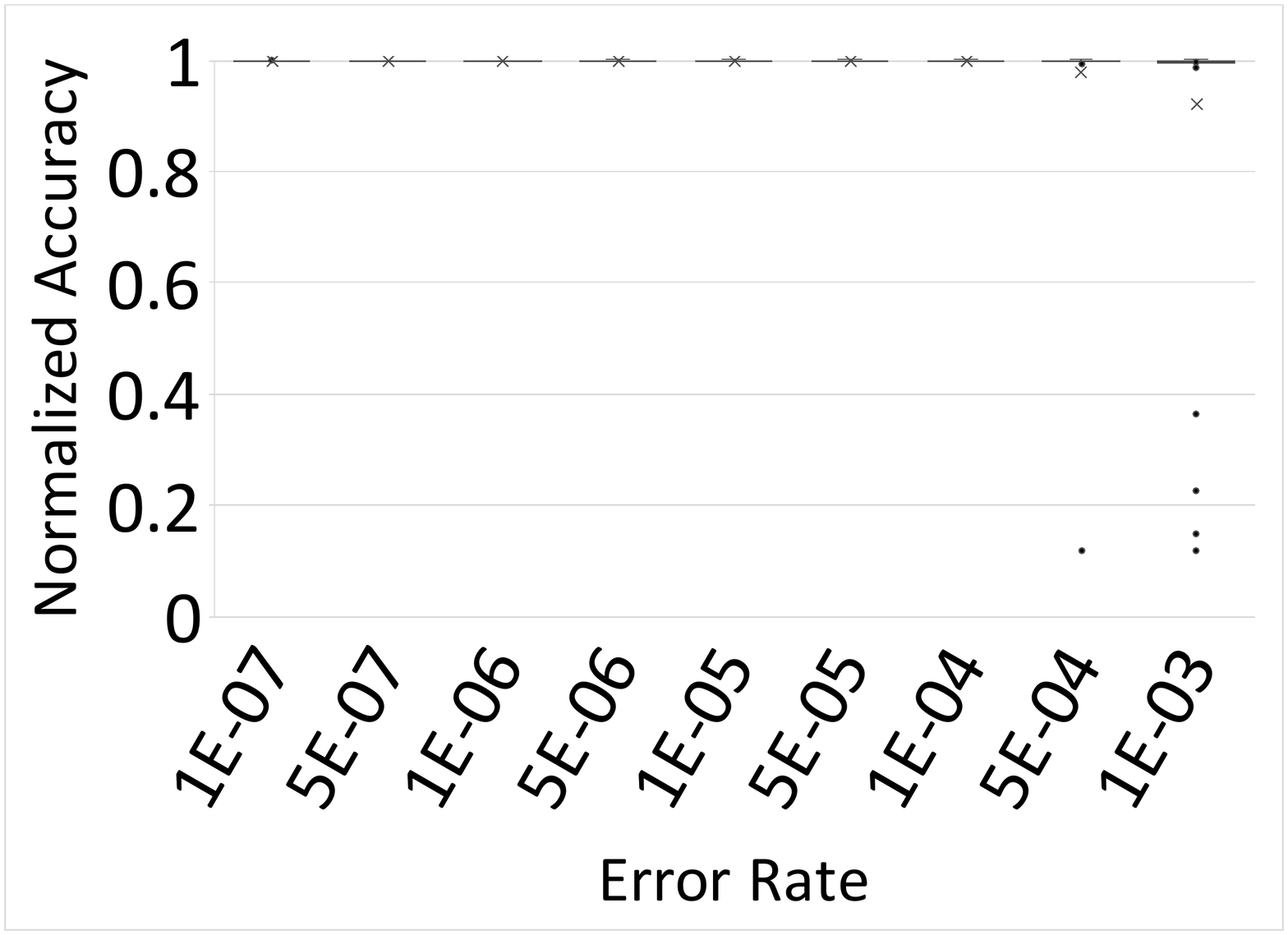}
            \label{fig:CIFAR_recov_whole}
        \end{minipage}
    }
    \caption{CIFAR-10 small network normalized accuracy after recovery from whole-weight errors}
    \label{fig:CIFAR-ACC-whole}
\end{figure}

When tested with whole-weight errors the network still had some intrinsic robustness, achieving \(100.0\%\) of accuracy up to an error rate of \(1E-04\). ECC was not tested as ECC's performance would match the no recovery performance.MILR performance did drop after this point as multiple erroneous layers between checkpoint caused performance degradation. This performance shows were MILR is most capable, large errors densely arranged, and where it beats out ECC's correcting ability. 
\begin{scriptsize}
    \begin{table}[htbp]
        \centering
        \caption{CIFAR-10 small network whole layer error accuracy }
        \label{tab:cifar_acc_wholeLayer}
        \begin{threeparttable}
            \begin{tabular}{|l|l|l|}
                \hline
                \textbf{Recovery} & \textbf{None} & \textbf{MILR} \\
                \hline
                \hline
                Conv &  12.0\% &  100.0\%  \\
                \hline
                Conv Bias &  63.3\% &  100.0\% \\
                \hline
                Conv 1 &  11.9\% &  N/A\tnote{*}  \\
                \hline
                Conv 1 Bias &  35.7\% & 100.0\% \\
                \hline
                Conv 2 &  11.8\% &  N/A\tnote{*} 	 \\
                \hline
                Conv 2 Bias &  80.7\% &  100.0\% \\
                \hline
                Conv 3 &  11.6\% &  N/A\tnote{*}  \\
                \hline
                Conv 3 Bias &  95.2\% & 100.0 \% \\
                \hline
                Conv 4 &  12.2\% &  NA\tnote{*}\\ 
                \hline
                Conv 4 Bias &  98.3\% &  100.0\% \\
                \hline
                Conv 5 &  11.8\% &  N/A\tnote{*}  \\
                \hline
                Conv 5 Bias &  98.9\% &  100.0\% \\
                \hline
                Conv 6 &  12.8\% &  N/A\tnote{*} \\
                \hline
                Conv 6 Bias &  98.6\% &  100.0\% \\
                \hline
                Dense &  11.8\% &  100.0\%  \\
                \hline
                Dense Bias &  99.8\% &  100.0\% \\
                \hline
                Dense 1 &  13.0\% &  100.0\% 	\\
                \hline
                Dense 1 Bias &  99.3\% &  100.0\% \\
                \hline
            \end{tabular}
            \begin{tablenotes}
                \item[*]Convolution partial recoverable
            \end{tablenotes}
        \end{threeparttable}
    \end{table}
\end{scriptsize}

The CIFAR-10 network with full layer errors had similar results to the MNIST network. It did have more partial recoverable layers as it has larger convolution layers. But MILR was capable of restoring all other layers to \(100.0\%\) of their original accuracy.
\begin{scriptsize}
    \begin{table}[htbp]
        \centering
        \caption{CIFAR-10 small network storage overhead }
        \label{tab:CIFAR_storage}
        \begin{tabular}{|l|l|l|l|}
            \hline
            \textbf{Backup Weights} & \textbf{ECC} & \textbf{MILR} & \textbf{ECC \& MILR} \\
            \hline
            \hline
            2.79 MB & 0.61 MB & 1.51 MB & 2.12 MB\\ 
            \hline
        \end{tabular}
    \end{table}
\end{scriptsize}

The small CIFAR-10 network's hyperparameters allowed for a lower storage overhead, shown in Table~\ref{tab:CIFAR_storage}, as it allowed for more data reuse. To store a second copy of the network it would cost 2.79 MB of storage. MILR only cost 1.51 MB of additional storage a \(45.9\%\) reduction in storage overhead. ECC is still cheaper only cost 0.61 MB, but still susceptible to multi-bit errors.  Combining ECC and MILR allows for all scenarios to be covered with single bit errors being handled by ECC and multi-bit errors being handled by MILR while costing less then storing a second copy of the network.

\subsection{CIFAR-10 Large Network}
The Cifar-10 dataset \cite{Krizhevsky2009LearningML} was used again with another model as the dataset is lightweight allowing for fast training and testing while representational of a real world use case. The large Cifar model is based off a model presented in the paper FAWCA \cite{FAWCA} and shown in Table~\ref{tab:cifar_B}, with same padding convolution layers; and bias and ReLu activation layer after each dense and convolution layer. This model is significantly larger then the small Cifar network, with larger and deeper filter along with larger dense layers. It was trained to \(83.6\%\) accuracy over 150 epochs with batch sizes of 128.
\begin{figure*}[htbp]
\scriptsize
\centering
    \subfloat[No recovery]{
        \begin{minipage}{0.235\textwidth}
            \centering
            \includegraphics[width=\textwidth]{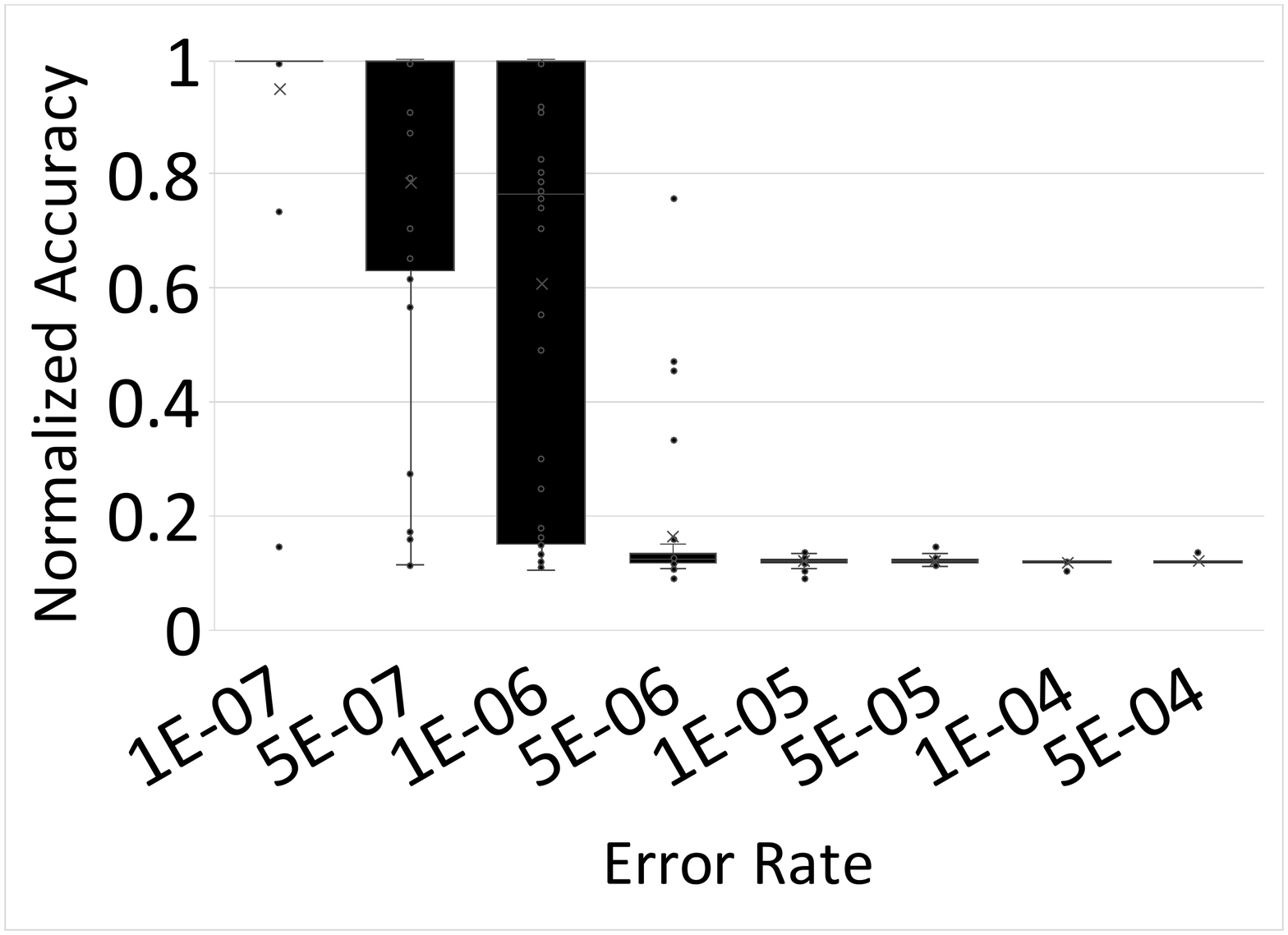}
            \label{fig:cifar_B_error}
        \end{minipage}
    }
    \subfloat[ECC]{
        \begin{minipage}{0.235\textwidth}
            \centering
            \includegraphics[width=\textwidth]{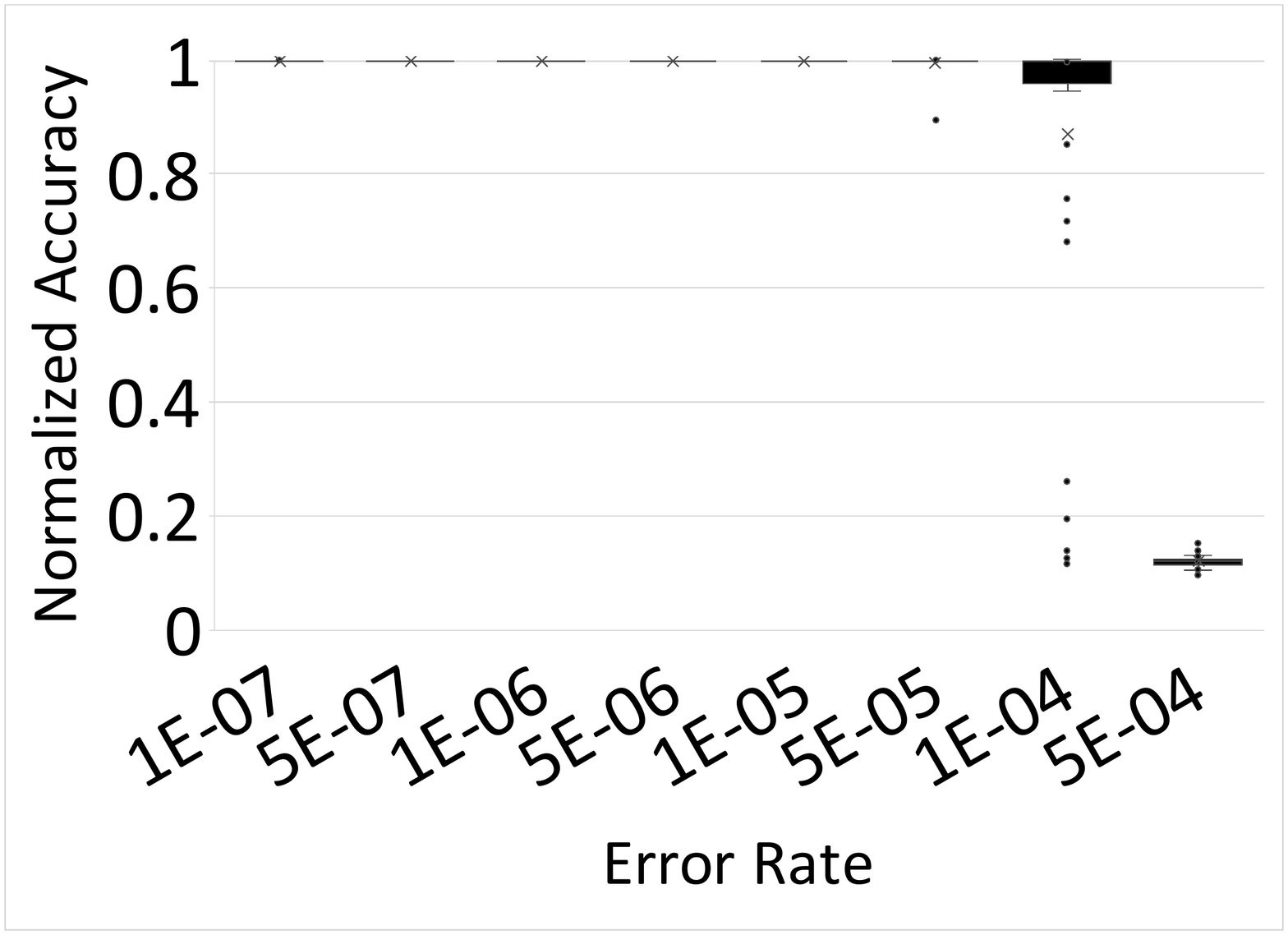}
            \label{fig:cifar_B_ecc}
        \end{minipage}
    } 
    \subfloat[MILR]{
        \begin{minipage}{0.235\textwidth}
            \centering
            \includegraphics[width=\textwidth]{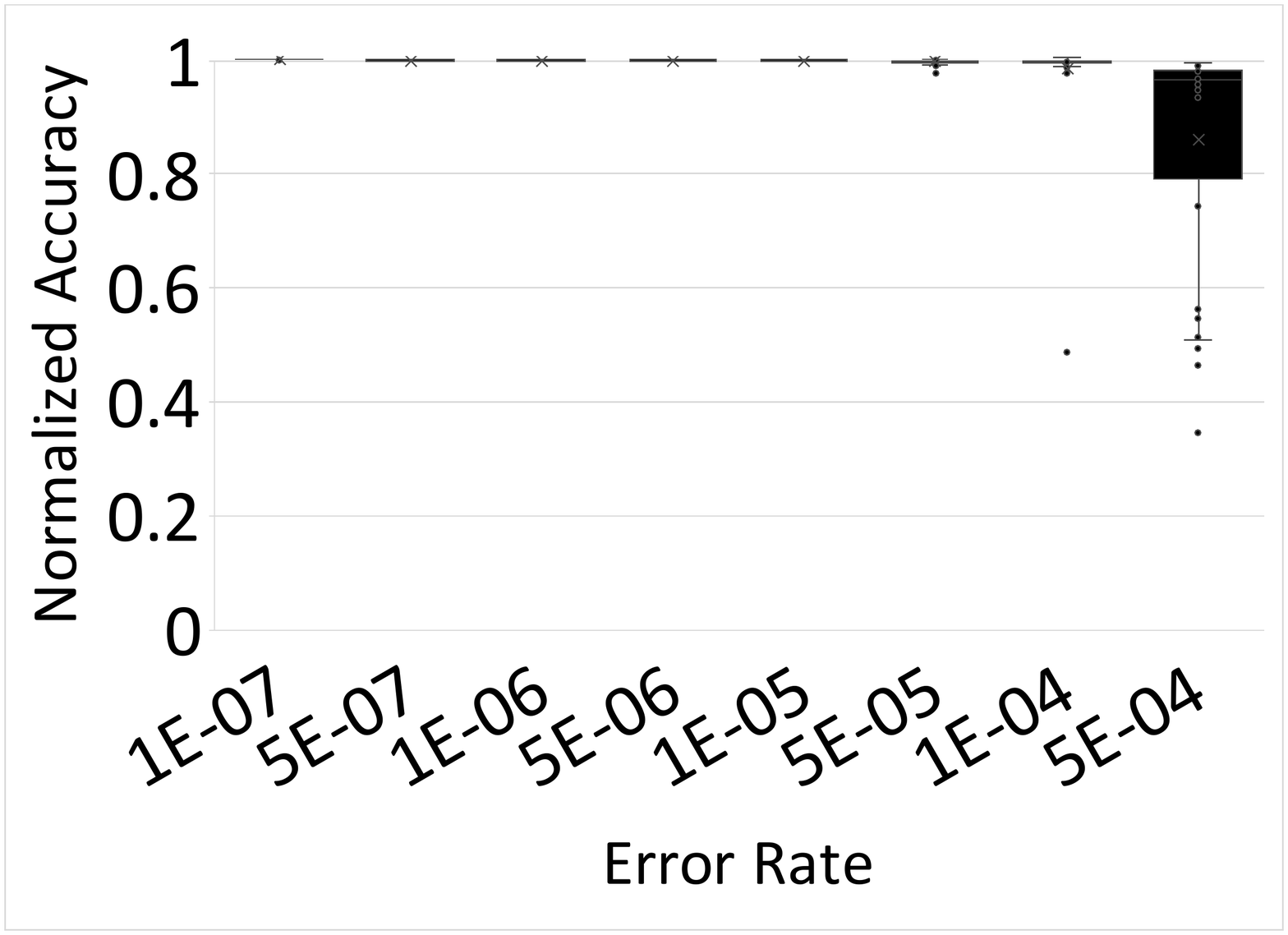}
            \label{fig:cifar_B_recov}
        \end{minipage}
    }
    \subfloat[ECC + MILR]{
        \begin{minipage}{0.235\textwidth}
            \centering
            \includegraphics[width=\textwidth]{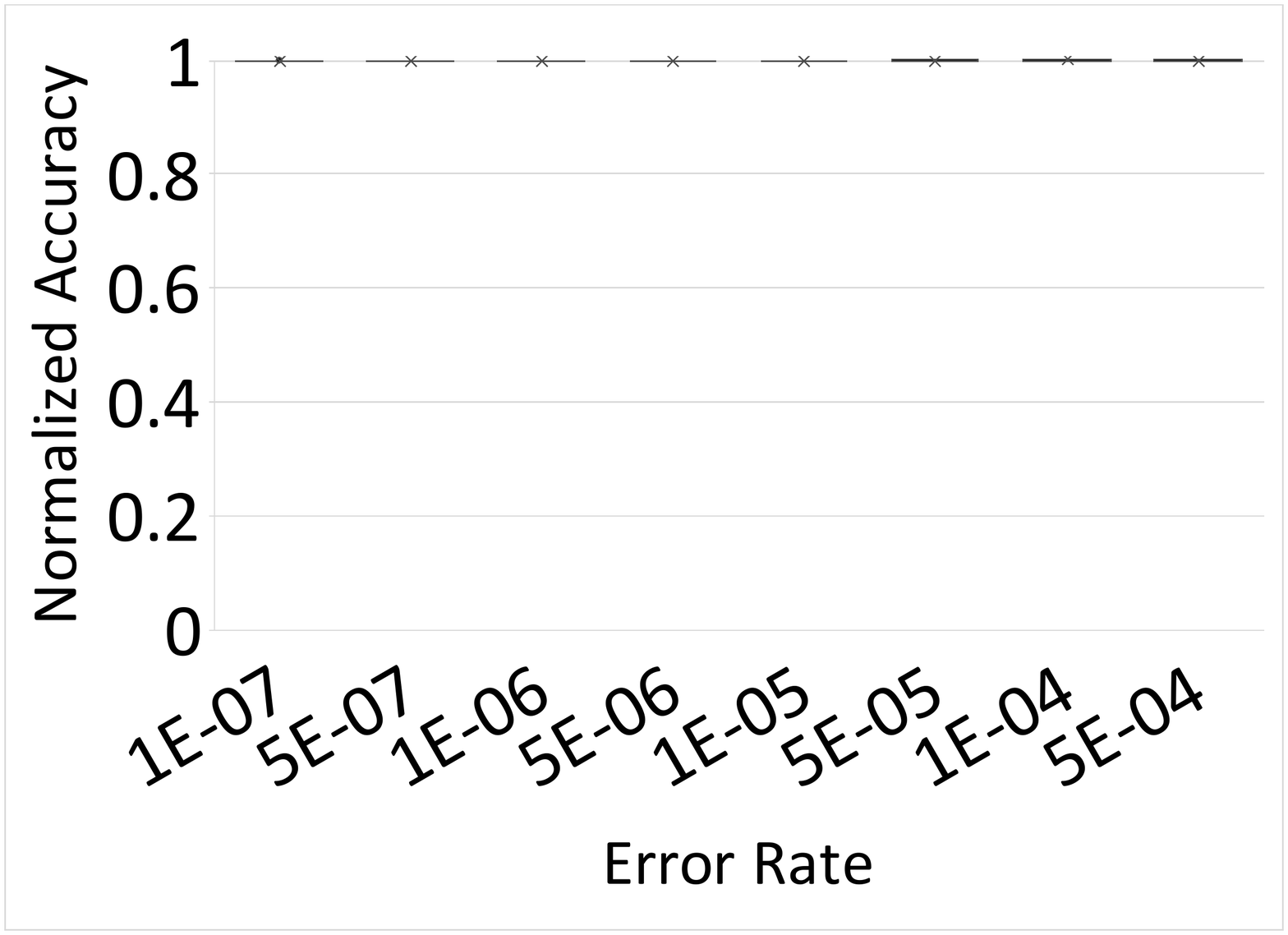}
            \label{fig:cifar_B_eccMilr}
        \end{minipage}
    }
    \caption{CIFAR-10 large network normalized accuracy after recovery from varying RBER}
    \label{fig:CIFAR-B_ACC}
\end{figure*}

The intrinsic robustness of the large cifar network is less than that of the small cifar, as due to the size even at the lower error there are significantly more errors. MILR is able to handle these additional errors as with the increase of size the number of errors MILR can recover is also increased naturally. MILR and ECC are both able to recover to \(100\%\) of accuracy through \(1E-05\) and start dropping after this point, but MILR is able to maintain an higher recoverabilty due to it not being affected by Multi-bit errors.
\begin{figure}[htbp]
    \scriptsize
    \centering
    \subfloat[No recovery]{
        \begin{minipage}{0.235\textwidth}
            \centering
            \includegraphics[width=\textwidth]{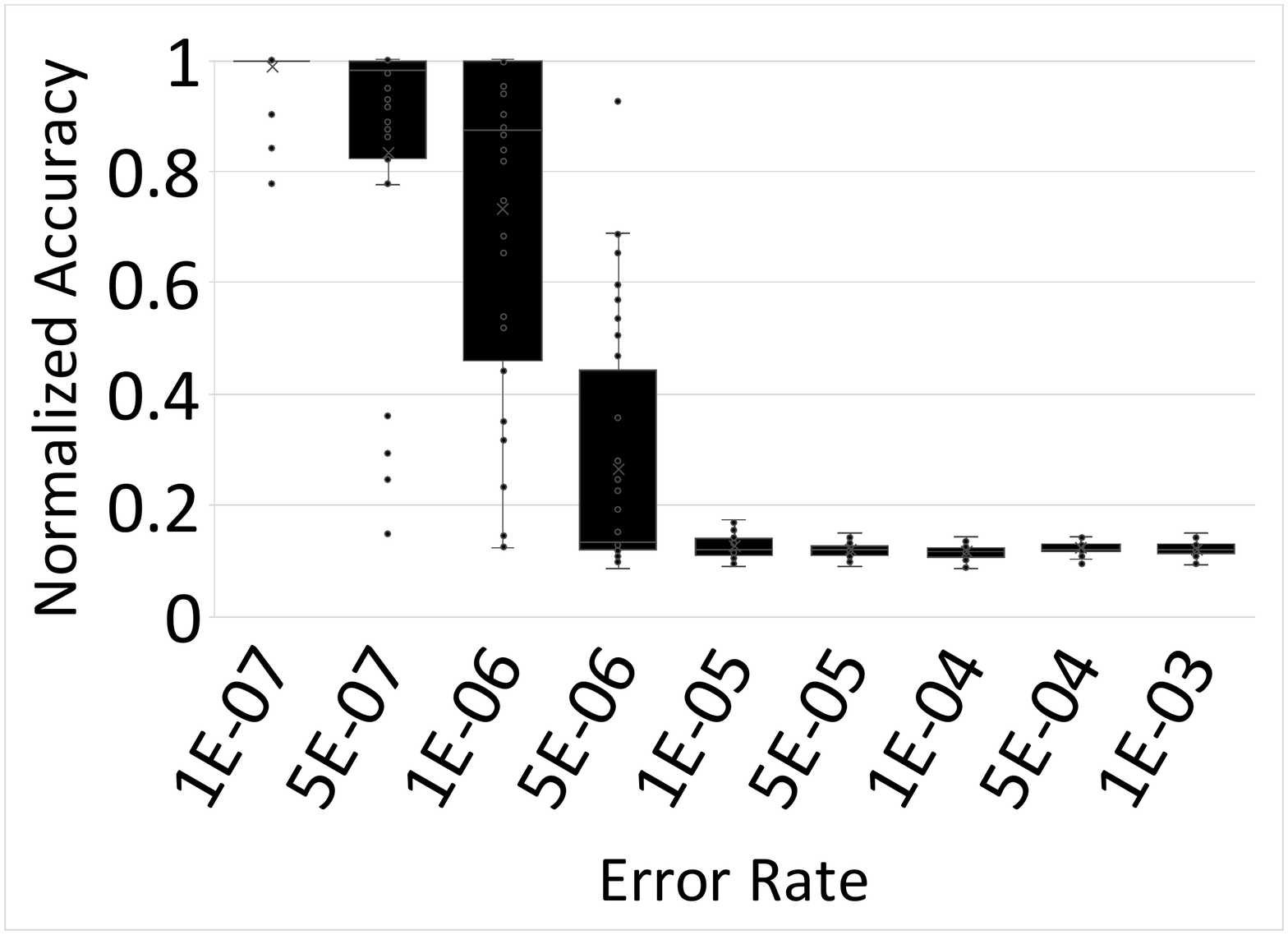}
            \label{fig:CIFAR_B_error_whole}
        \end{minipage}
    }
    \subfloat[MILR]{
        \begin{minipage}{0.235\textwidth}
            \centering
            \includegraphics[width=\textwidth]{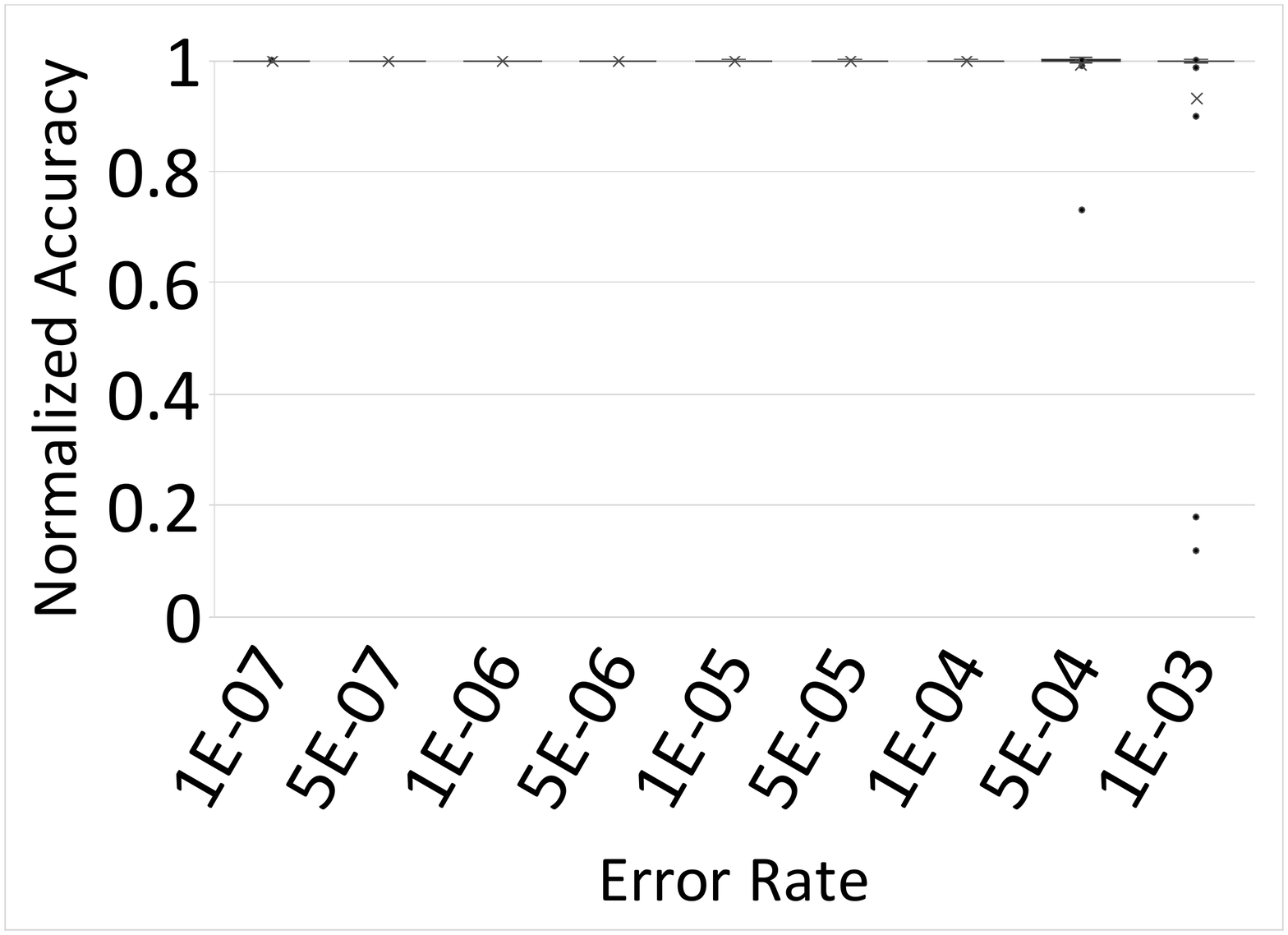}
            \label{fig:CIFAR_B_recov_whole}
        \end{minipage}
    }
    \caption{CIFAR-10 large network normalized accuracy after recovery from whole-weight errors}
    \label{fig:CIFAR-B-ACC-whole}
\end{figure}

For whole word error, Figure~\ref{fig:CIFAR-B-ACC-whole} the results are also similar to the small cifar network. MILR is able to achieve high recoverabilty until multiple erroneous layers occur between checkpoints. 
\begin{scriptsize}
    \begin{table}[htbp]
        \centering
        \caption{CIFAR-10 large network whole layer error accuracy }
        \label{tab:cifar_B_acc_wholeLayer}
        \begin{threeparttable}
            \begin{tabular}{|l|l|l|}
                \hline
                \textbf{Recovery} & \textbf{None} & \textbf{MILR} \\
                \hline
                \hline
                Conv &  12.2\% &   N/A\tnote{*}  \\
                \hline
                Conv Bias &  24.4\% &  100.0\% \\
                \hline
                Conv 1 &  9.6\% &  N/A\tnote{*}  \\
                \hline
                Conv 1 Bias &  71.5\% & 100.0\% \\
                \hline
                Conv 2 &  11.8\% &  N/A\tnote{*} 	 \\
                \hline
                Conv 2 Bias &  85.6\% &  100.0\% \\
                \hline
                Conv 3 &  12.4\% &  N/A\tnote{*}  \\
                \hline
                Conv 3 Bias &  95.8\% & 100.0 \% \\
                \hline
                Conv 4 &  12.0\% &  NA\tnote{*}\\ 
                \hline
                Conv 4 Bias &  97.1\% &  100.0\% \\
                \hline
                Conv 5 &  11.4\% &  N/A\tnote{*}  \\
                \hline
                Conv 5 Bias &  98.9\% &  100.0\% \\
                \hline
                Dense &  12.0\% &  100.0\%  \\
                \hline
                Dense Bias &  100.0\% &  100.0\% \\
                \hline
                Dense 1 &  12.1\% &  100.0\% 	\\
                \hline
                Dense 1 Bias &  99.7\% &  100.0\% \\
                \hline
            \end{tabular}
            \begin{tablenotes}
                \item[*]Convolution partial recoverable
            \end{tablenotes}
        \end{threeparttable}
    \end{table}
\end{scriptsize}

With the larger size of this network the convolution layers were required to use partial recoverability to keep cost low. This means none of the convolution layers are able to cope with being completely modified, they are limited to \(G^2\) erroneous parameters per filter. For the rest of the layers MILR was able to recover them back to \(100.0\%\) of their original accuracy as shown in Table~\ref{tab:cifar_B_acc_wholeLayer}.
\begin{scriptsize}
    \begin{table}[htbp]
        \centering
        \caption{CIFAR-10 large network storage overhead }
        \label{tab:CIFAR_B_storage}
        \begin{tabular}{|l|l|l|l|}
            \hline
            \textbf{Backup Weights} & \textbf{ECC} & \textbf{MILR} & \textbf{ECC \& MILR} \\
            \hline
            \hline
            9.56 MB & 2.09 MB & 8.50 MB & 9.59 MB\\ 
            \hline
        \end{tabular}
    \end{table}
\end{scriptsize}

The cost of the large Cifar network cost more than the small network due to the larger size. However was able to keep cost lower than storing a second copy of the network thanks to the user of convolution layer partial recoverabilty. MILR came in with a cost of 8.50 MB, a \(11.0\%\) reduction in cost over storing a backup copy of the network.

\subsection{Network Availability}
The availability of the CNN has a major impact on the usefulness of the network, but availability and accuracy can be an important trade off in a CNN system. This is due to availability being reduced when a network has to recover from errors, and without recovery a networks accuracy can degrade. Therefore systems have to find a balance that suits their intended mission. In a mission critical application, such as a self driving car, the need high accuracy might lessen the need for availability as redundancy already exists. If a network has a high availability requirement, such as a website recommendation tool, accuracy might not be as important but it always need to be available for the user.

\begin{equation}
\label{eq:availabilty}
   f(a) = A\left(\frac{1}{\cfrac{\frac{1}{a}-1}{\cfrac{(T_{d}I)+T_{r}}{T_{be}}}}\right)
\end{equation}

This trade off can be modeled by using equation \ref{eq:availabilty}. Where \(a\) is the required availability, \(T_{d}\) is the time taken in the detection phase, \(I\) is the number of runs of detection between errors, \(T_{r}\) is the time taken to recover, \(T_{be}\) is the time between errors in the system, and there exist a function \(A()\) that given the number of errors returns the network accuracy. 

We evaluated performance and availability on a MILR system running on Windows 10 OS on a Ryzen 5 2600X, 32 GB of RAM and a Nvidia RTX 2070.  MILR was able to take advantage of the GPU for parts of the layer solving, but much of the operations were confined to the CPU. There is still potential optimization of MILR that can improve the performance, but MILR was evaluated in its current state. 
\begin{scriptsize}
    \begin{table}[htbp]
        \centering
        \caption{MILR prediction and identification time in seconds }
        \label{tab:Ident_time}
        \begin{tabular}{|l|l|l|l|}
            \hline
            \textbf{Network} & \textbf{Single } & \textbf{Batch} & \textbf{Identification}  \\
             & \textbf{Prediction} & \textbf{Prediction} &  \\
            \hline
            \hline
            MNIST & 0.017s & 3.48E-05s& 0.010s \\
            \hline
            CIFAR-10 Small & 0.018s &6.50E-05s& 0.018s \\
            \hline
            CIFAR-10 Large & 0.018s &8.77E-05s& 0.016s \\
            \hline
        \end{tabular}
    \end{table}
\end{scriptsize}

Error identification time varies between networks but stays constant in each network, with times shown in Table~\ref{tab:Ident_time}. When compared to a single prediction, the times are comparable and reasonable as MILR uses a forward pass in its error detection. Compared to a prediction run in a large batches the performance can be 200\(\times\)-300\(\times\) slower as batch operations are able to take advantage of the pipelining of the predictions. 
\begin{figure}[htbp]
\begin{tikzpicture}
\begin{axis}[
    axis lines = left,
    xlabel = Errors (thousands),
    ylabel = Recovery Time (s),
    xmin=-0, xmax=10,
    ymin=0, ymax=5
]
\addplot [
    color=red,
]coordinates {
        (0,1.28250001125707E-06)
        (0,1.28249996578233E-06)
        (0,1.39499998681458E-06)
        (0,1.30250000438536E-06)
        (0.00025,0.0126899374999993)
        (0.00375,0.125825402499998)
        (0.00575,0.150894352500017)
        (0.0158,0.173678990000007)
        (0.026225,0.173866457500002)
        (0.051025,0.182307282500033)
        (0.2596,0.201592180000011)
        (0.3882,0.197377315000002)
        (0.520825,0.211338915000021)
        (1.52565,0.2312915825)
        (2.5944,0.262749600000023)
        (5.1691,0.313640829999985)
        (25.89185,1.16205594249999)
        (51.73905,3.84934523)
	};
\addlegendentry{MNIST}
\addplot [
    color=blue,
    ]coordinates {
		(0,1.52499987962073E-06)
        (0,1.49249992773548E-06)
        (0,1.59250005253852E-06)
        (0,1.58250004460569E-06)
        (0.00005,0.0041770325000698)
        (0.00185,0.0573701375001291)
        (0.00195,0.0916284875000199)
        (0.006575,0.186011339999936)
        (0.010825,0.321522899999991)
        (0.02145,0.452580069999942)
        (0.11015,0.593283780000036)
        (0.15785,0.574255787499919)
        (0.21875,0.675570924999965)
        (1.088875,1.3813435575)
        (2.1598,2.70805964000002)
        (10.833075,37.8472290175)
	};
\addlegendentry{CIFAR-10 Small}

\addplot [
    color=green,
    ]
    coordinates {
		(0,1.4050000345378E-06)
        (0,1.72249999650375E-06)
        (0,1.77000001713167E-06)
        (0,1.41749994782003E-06)
        (0.00025,0.0149463699999842)
        (0.005,0.199981132499988)
        (0.0075,0.303921909999997)
        (0.0214,0.558546682500128)
        (0.0367,0.786199239999996)
        (0.0736,0.966941860000018)
        (0.36825,1.18759049499999)
        (0.555775,1.14246339499995)
        (0.7396,1.33306778500001)
        (3.70225,3.0700002)
        (7.4012,7.038936595)
        (37.006625,138.684157112499)
	};
\addlegendentry{CIFAR-10 Large}

\end{axis}
\end{tikzpicture}
\caption{The relation between recovery time and errors }
\label{fig:milr-timegraph}
\end{figure}
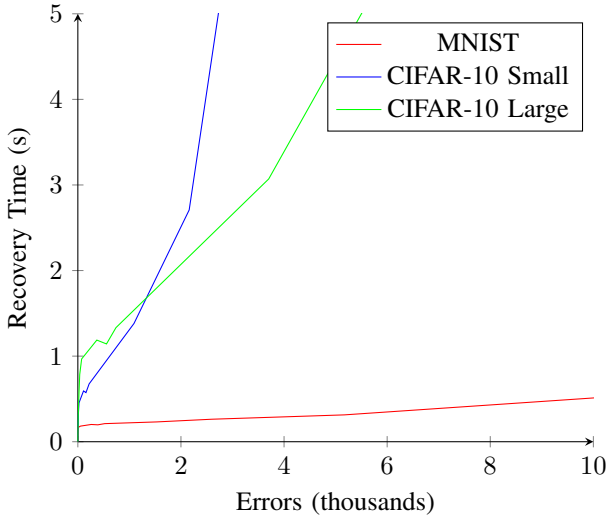

Error recovery times are dependent on the number of errors as shown in Figure~\ref{fig:milr-timegraph}. As the number of errors grow, the recovery time also grows due to solving for more errors in the partial recoverability of convolution layers, plus  additional layers needing solving. The growth rate is unique to each network, and also increases super linearly. These time cost can be balanced by inducing recovery before the number of errors exceed a point in which the recovery time increases exponentially.
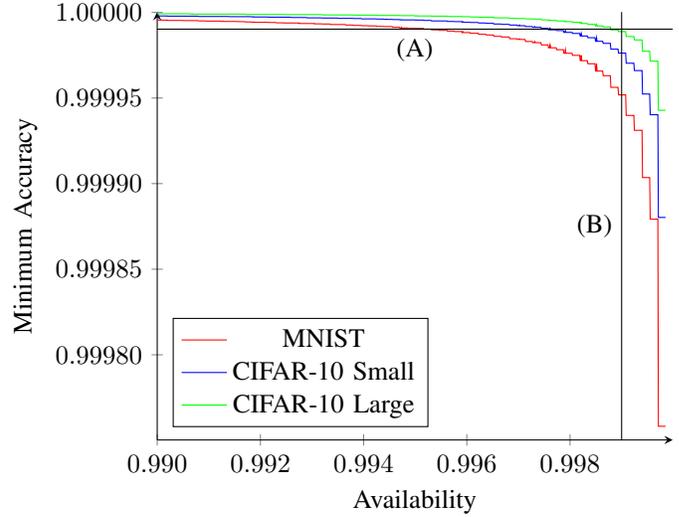
\begin{figure}[htbp]
\begin{tikzpicture}
\begin{axis}[
    axis lines = left,
    xlabel = Availability,
    ylabel = Minimum Accuracy,
    y tick label style={
        /pgf/number format/.cd,
            fixed,
            fixed zerofill,
            precision=5,
        /tikz/.cd
    },
    x tick label style={
        /pgf/number format/.cd,
            fixed,
            fixed zerofill,
            precision=3,
        /tikz/.cd
    },
    legend pos=south west,
]
\addplot [
    domain=0.99:.9999, 
    samples=1000, 
    color=red,
]
{(1/((1/x)-1)/((0.0100 * 2)+0.182)/898203.5928
)* -0.008782349 )+1};
\addlegendentry{MNIST}

\addplot [
     domain=.99:.9999, 
    samples=1000, 
    color=blue,
    ]
    {(1/((1/x)-1)/((0.0180 * 2)+0.453)/2150537.634)* -0.025228571 )+1};
\addlegendentry{CIFAR-10 Small}

\addplot [
     domain=.99:.9999, 
    samples=1000, 
    color=green,
    ]
    {(1/((1/x)-1)/((0.0160 * 2)+0.97)/627615.0628)* -0.007230896 )+1};
\addlegendentry{CIFAR-10 Large}

\addplot [
    domain=0.99:.99999, 
    color=black,
    ] coordinates {(.999, 0.99975) (.999, 1)} node[left,pos=.5] {(B)};
\addplot [
    domain=0.99:.99999, 
    color=black,
    ]{.99999} node[below,pos=.5] {(A)};

\end{axis}
\end{tikzpicture}
\caption{The trade off between availability and minimum accuracy, (A) Minimum Accuracy of \(99.999\%\), (B) Availability of \(99.9\%\)}
\label{fig:availabilty}
\end{figure}

    
    

Using MILR's identification and recovery time this balance between accuracy and availability is plotted in Figure~\ref{fig:availabilty}. Assuming a worst case mean time between failures of 75,000 errors per billion device hours per Mbit~\cite{Schroeder2009DRAMEI}, where each bit error affected a ciphertext word causing multi-bit errors int the plaintext, error detection runs twice between error intervals, recovery time is the maximum recovery time for the expected errors in a single year, and that an accuracy equation \(A(n)\) exist and is linear degradation of accuracy from zero errors and the expected errors in a single year.

Figure~\ref{fig:availabilty} show that their is a trade off between accuracy and availability. This graph is useful to determine the settings of error detection intervals in MILR. Two example users (A and B) are shown in the graph. User A needs a high accuracy network that sustained at least \(99.999\%\) accuracy, the availability that each network yields is shown in the intersection of line (A) and the networks. On the other hand, user B needs availability of at least 99.9\%, and the obtained accuracy for each network is shown by the intersection of line (B) and the networks. 

\section{Conclusion}
In this paper we made a novel distinction between ciphertext-space vs. plaintext-space error correction (PSEC). We pointed out that the assumption of randomly distributed bit errors ECC relies on is not valid in plaintext space, where one bit error in the ciphertext space turns into concentrated many-bit errors affecting encryption words, causing whole-weight errors that are difficult to recover from using ECC. We then introduced MILR, to our knowledge the first PSEC  technique for CNNs. MILR takes advantage of the natural algebraic relationship between input, parameters, and output of CNNs, in order to detect and correct bit errors, whole-weight errors, and even whole-layer errors.  MILR is implemented in software and can run on any hardware. MILR can detect and correct errors which is difficult to achieve using ECC, unlocking robust PSEC for situations where CNNs run on encrypted VM. We demonstrate that MILR can recover from whole-weight errors even at up to 1E-03 error rate, while even  whole-layer errors can even be recovered to 100\% accuracy, making MILR a suitable choice for PSEC. Even on random bit errors, MILR outperforms ECC in keeping network accuracy high.



\bibliographystyle{IEEEtranS}
\bibliography{Sections/references}

\end{document}